\documentclass[10pt,twocolumn,letterpaper]{article}

\usepackage[pagenumbers]{cvpr} 

\definecolor{cvprblue}{rgb}{0.21,0.49,0.74}
\usepackage[pagebackref,breaklinks,colorlinks,allcolors=cvprblue]{hyperref}
\usepackage{multirow}
\usepackage{graphicx}
\usepackage{float}

\title{3DAffordSplat: Efficient Affordance Reasoning with 3D Gaussians}

\author{
Zeming Wei$^1$\thanks{Equal contribution} \hspace{0.7em} Junyi Lin$^{1*}$ \hspace{0.7em} Yang Liu$^{1,3}$\thanks{Corresponding Author}
\hspace{0.7em} Weixing Chen$^1$
\hspace{0.7em} Jingzhou Luo$^1$
\hspace{0.7em}  Guanbin Li$^{1,2,3}$\\ \hspace{0.7em}  Liang Lin$^{1,2,3}$\\
$^1$Sun Yat-sen University, China $^2$Peng Cheng Laboratory \\$^3$Guangdong Key Laboratory of Big Data Analysis and Processing\\
{\tt\small \{weizem6,linjy279\}@mail2.sysu.edu.cn,liuy856@mail.sysu.edu.cn,\{chenwx228,luojzh5\}@gmail.com}\\
{\tt\small liguanbin@mail.sysu.edu.cn,linliang@ieee.org}\\
{\tt\small \href{https://github.com/HCPLab-SYSU/3DAffordSplat}{github.com/HCPLab-SYSU/3DAffordSplat}}
}

\begin{document}
\maketitle
\begin{abstract}
3D affordance reasoning is essential in associating human instructions with the functional regions of 3D objects, facilitating precise, task-oriented manipulations in embodied AI. However, current methods, which predominantly depend on sparse 3D point clouds, exhibit limited generalizability and robustness due to their sensitivity to coordinate variations and the inherent sparsity of the data. By contrast, 3D Gaussian Splatting (3DGS) delivers high-fidelity, real-time rendering with minimal computational overhead by representing scenes as dense, continuous distributions. This positions 3DGS as a highly effective approach for capturing fine-grained affordance details and improving recognition accuracy. Nevertheless, its full potential remains largely untapped due to the absence of large-scale, 3DGS-specific affordance datasets.
To overcome these limitations, we present \textbf{3DAffordSplat}, the first large-scale, multi-modal dataset tailored for 3DGS-based affordance reasoning. This dataset includes 23,677 Gaussian instances, 8,354 point cloud instances, and 6,631 manually annotated affordance labels, encompassing 21 object categories and 18 affordance types. Building upon this dataset, we introduce \textbf{AffordSplatNet}, a novel model specifically designed for affordance reasoning using 3DGS representations. AffordSplatNet features an innovative cross-modal structure alignment module that exploits structural consistency priors to align 3D point cloud and 3DGS representations, resulting in enhanced affordance recognition accuracy. Extensive experiments demonstrate that the 3DAffordSplat dataset significantly advances affordance learning within the 3DGS domain, while AffordSplatNet consistently outperforms existing methods across both seen and unseen settings, highlighting its robust generalization capabilities.

\end{abstract}    
\section{Introduction}
\label{sec:intro}

3D affordance reasoning represents a fundamental capability for embodied agents to understand how to interact with objects in their environment ~\cite{gibson1977theory,liu2024aligning,ren2024infiniteworld}. 
By identifying functional regions of 3D objects that allow specific actions (e.g., parts that can be grasped, pulled, or rotated), robots can perform precise manipulations based on human instructions ~\cite{wu2023learning,tang2023towards,liu2023cross,DBLP:conf/corl/JiQZW24,nasiriany2024rtaffordance,shorinwa2024splatmover,wei2025afforddexgrasp,zhu2025afford,car2024plato,chen2025cross}. 
This capability bridges the gap between perception and action, enabling more natural human-robot collaboration in various applications ranging from household assistance to industrial automation.

\begin{figure}[t]
    \centering
\includegraphics[width=0.48\textwidth]{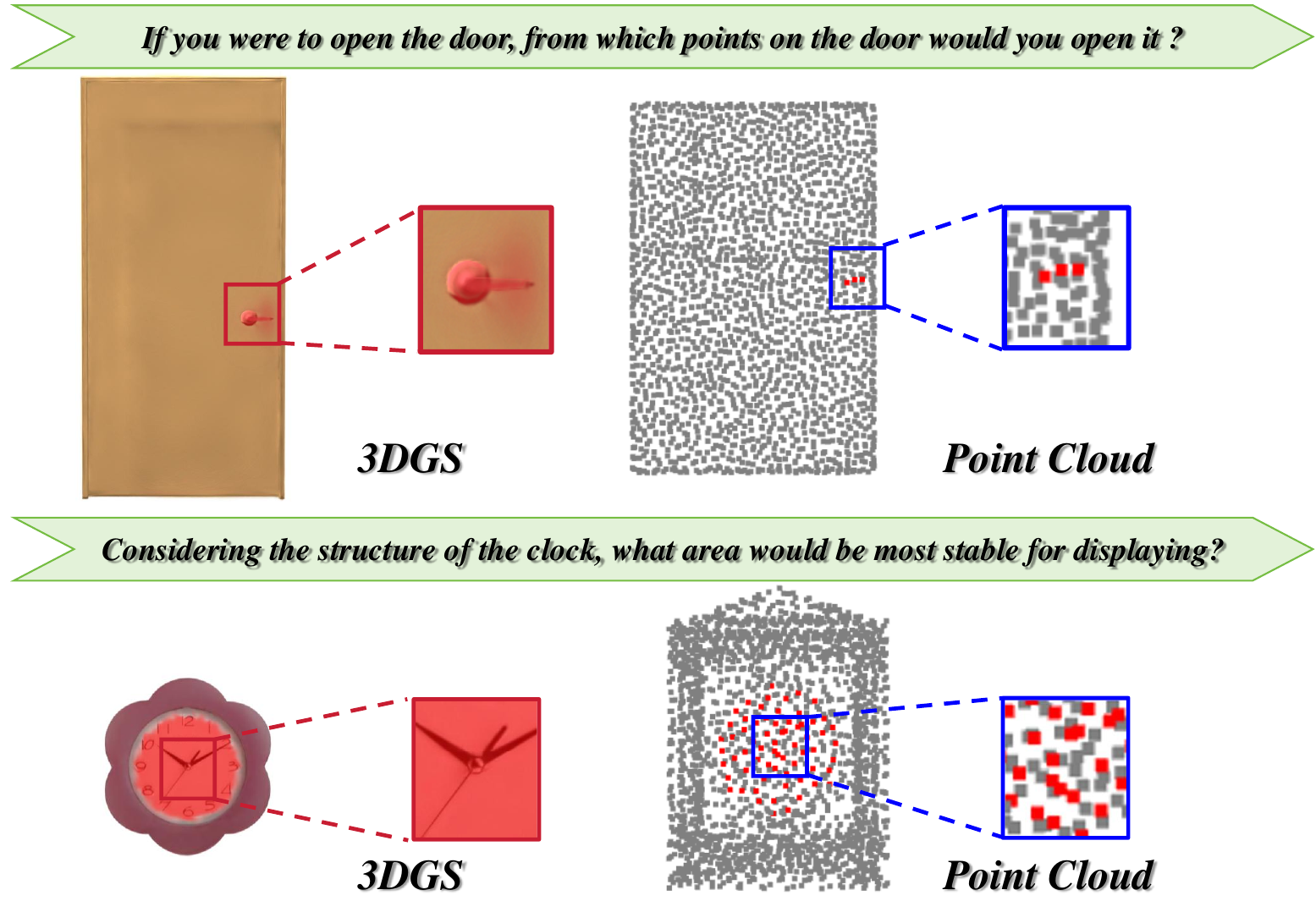}
    \caption{Compared to sparse point clouds, 3DGS provides more vivid textures and clearer geometry. 3DGS-based Affordances can capture more complex structures. Moreover, the continuous nature of Gaussians supports smooth affordance representation over surfaces and even curves.}
\label{fig:Experiment1}
\end{figure}


Existing methods for affordance reasoning primarily rely on image, video, and point cloud representations~\cite{li2024laso,liu2019deep,shao2024greatgeometryintentioncollaborativeinference, bahl2023affordances,luo2025dspnet}. However, each of these approaches presents notable limitations. Image-based methods depend solely on 2D projections, which lack depth information and fail to capture the complete 3D structure of objects~\cite{yan2023skeletonmae,li2024one}. While videos provide dynamic visual cues, they do not offer direct 3D spatial information and are challenging to annotate~\cite{bahl2023affordances}. Additionally, videos often struggle to represent subtle dynamic changes during human-object interactions. Point cloud data, although providing direct 3D geometric representation, are inherently discrete~\cite{li2024laso, lu2024geal, yang2023grounding, gao2024learning, yu2024seqafford,jiang2025beyond}. As shown in \autoref{fig:Experiment1}, their sparsity and limited geometric resolution fundamentally constrain their ability to represent detailed and continuous affordance structures. This critical limitation arises from their discrete sampling nature, which fails to capture continuous surfaces and intricate geometric features essential for precise reasoning by AI agents. 

Recent advances in 3D Gaussian Splatting (3DGS)~\cite{kerbl20233d} offer promising solutions, enabling high-fidelity scene reconstruction and real-time rendering through Gaussian primitives that inherently encode rich 3D geometric and photometric attributes. 3DGS represents 3D scenes as a collection of 3D Gaussians with learnable parameters, offering several advantages over traditional 3D Affordance approach: 
1) higher geometric precision and the preservation of surface details, addressing the issues of discreteness and incompleteness in point cloud data, 
2) integration of rich color information, compensating for the lack of 3D spatial information in image-based methods, 
3) efficient real-time rendering with low computational requirements, achieving high frame rates (30+ fps at 1080p resolution) and overcoming the limitations of video-based methods in dynamic information capture and resource efficiency. 
These properties make 3DGS particularly suitable for affordance reasoning in embodied intelligence applications where real-time performance and resource efficiency are critical.

Despite the advantages of 3DGS, its application in affordance reasoning is hindered by three significant challenges. The lack of large-scale 3DGS datasets with affordance annotations limits model training and evaluation, while existing models, designed for discrete data like point clouds or images, fail to leverage 3DGS’s unique continuous properties, reducing potential gains in accuracy and efficiency. Additionally, aligning 3DGS with abundant point cloud affordance data is complex due to the mismatch between point clouds’ sparse, noisy nature and 3DGS’s detailed, continuous representation, requiring elaborated techniques to ensure geometric and semantic consistency. More importantly, conventional semantic embedding methods for 3DGS suffer from fundamental limitations~\cite{choi2024click, DBLP:journals/corr/abs-2312-00860, DBLP:conf/corl/JiQZW24}. Parametric expansion techniques that statically assign a single semantic feature to each Gaussian primitive are inadequate for representing multi-attribute affordance scenarios, in which individual Gaussian may simultaneously contribute to diverse functional contexts. This constraint on single semantics reduces real-world applicability, as objects often require context-aware interpretations across multiple affordance dimensions.


To address these challenges, we first introduce \textbf{3DAffordSplat}, the first large-scale, multi-modal 3DGS-based Affordance Reasoning dataset with comprehensive affordance annotations. As shown in \autoref{fig:teaser}, 3DAffordSplat encompasses three modalities: 3D Gaussian, point cloud, and textual instruction, all aligned with consistent affordance annotations.  This dataset supports effective cross-modal learning and facilitates knowledge transfer across various representations. Furthermore, 3DAffordSplat comprises a diverse array of objects and scenes, providing a robust foundation for developing and evaluating affordance reasoning models. 

Building on this dataset, we establish the first comprehensive evaluation framework for 3DGS-based affordance reasoning. Our benchmark employs established metrics from prior affordance analysis research~\cite{li2024laso, yang2023grounding} - including mIoU, AUC, SIM and MAE - to enable cross-modal performance comparison while maintaining backward compatibility with existing point cloud benchmarks. This framework facilitates fair comparisons between different methods and provides a new direction for advancing research in this domain.

Additionally, we propose a novel 3DGS-based affordance reasoning model, \textbf{AffordSplatNet}, the first generalizable 3DGS architecture for affordance reasoning that establishes cross-modal structural correspondence between sparse point clouds and dense Gaussian representations.
Our model incorporates a cross-modal structure alignment module that utilizes structural consistency priors to align 3D point cloud and 3DGS representations. This effective alignment and knowledge transfer between complementary representations not only enhances affordance reasoning precision but also improves the  robustness to geometric variations and partial observations. Our contributions are summarized as follows.
\begin{itemize}
\item We introduce 3DAffordSplat, the first large-scale, multi-modal 3DGS-based Affordance Reasoning with comprehensive affordance annotations, comprising Gaussian, point cloud, and textual instruction modalities.


\item We propose a novel 3DGS-based affordance reasoning model, AffordSplatNet, that enables effective knowledge transfer between point cloud and Gaussian representations, improving affordance reasoning accuracy and robustness.

\item Extensive experiments demonstrate that  3DAffordSplat effectively enhances existing point cloud methods for 3DGS affordance reasoning. Additionally, our AffordSplatNet outperforms existing methods in both seen and unseen settings, validating its generalization ability.

\end{itemize}


\begin{table*}[t]\small
    \caption{Comparison with existing 3D Affordance datasets. 3DAffordSplat uniquely integrates 3DGS, point clouds, and language. It contains 8.4k point clouds, 23k 3DGS, and 6,631 fine-grained 3DGS affordance annotations. ``Reasoning'' involves language-guided affordance recognition and text response generation, ``Grounding'' focuses solely on affordance region output, and ``No limit'' indicates that this dataset serves as a general-purpose dataset without specific restrictions.}
    \centering
        \setlength{\tabcolsep}{3pt}
    \begin{tabular}{c|ccccccc}
    \toprule
        \multirow{2}{*}{\textbf{Benchmark}} & \multirow{2}{*}{\textbf{Research Subject}} & \multicolumn{4}{c}{\textbf{Components}} & \multirow{2}{*}{\textbf{3DGS Affordance Annotations}} & \multirow{2}{*}{\textbf{Task Type}}\\
        \cline{3-6}
          & & 3D Gaussians & Point Clouds & Text & Image & & \\
        \midrule
                3DAffordanceNet~\cite{deng20213d}&  Point Clouds&  none& 56k &  $\times$& $\times$& none& No limit\\
                PIAD~\cite{yang2023grounding}&  Point Clouds&  none& 7k &  $\times$& \checkmark& none&  Grounding\\
                LASO~\cite{li2024laso}&  Point Clouds&  none& 8.4k &  \checkmark& $\times$& none& Reasoning\\
                PIAD-C~\cite{lu2024geal}& Point Clouds&  none& 2.5k &  $\times$& \checkmark& none&  Grounding\\
                LASO-C~\cite{lu2024geal}&  Point Clouds&  none& 2.4k &  \checkmark& $\times$& none& Reasoning\\
                PIADv2~\cite{shao2024greatgeometryintentioncollaborativeinference}&  Point Clouds&  none& 38k &  $\times$& \checkmark& none&  Grounding\\
                SeqAfford~\cite{yu2024seqafford}&  Point Clouds&  none& 1.8k &  \checkmark& $\times$& none&  Reasoning\\
                AGPIL~\cite{zhu2025grounding3dobjectaffordance}&  Point Clouds&  none& 41k &  \checkmark& \checkmark& none&  Reasoning\\
                \textbf{3DAffordSplat (Ours)}&  \textbf{3D Gaussians}&  \textbf{23k}& 8.4k &  \checkmark& $\times$& \textbf{6,631}& Reasoning\\
    \bottomrule
\end{tabular}\label{tab:DatasetComparison}
\end{table*}

\section{Related Work}
\label{sec:related work}
\subsection{Affordance Learning}

Initial efforts in affordance learning first concentrated on 2D domain. Early methods~\cite{do2018affordancenet} mainly focused on locating interaction regions in images and videos and then grouding~\cite{luo2022learning,bahl2023affordances,li2023locate,zhao2020object} the affordance.
These works relied mainly on precise annotations and convolutional neural networks (CNNs). 
To address the limitation of semantics and dynamic granularity, some researchers~\cite{li2024one,jang2024intra} incorporated language with 2D images.
Latest 2D work focused on limited sample~\cite{li2024one}, the combination of large language models (LLMs)~\cite{qian2024affordancellm} and embodied learning~\cite{zhu2025afford,heidinger20252handedafforder}, to cut down the cost and embracing the real world. 
However, 2D domain leads to some fatal problem. On one hand, there is a limitation on complex 3D interactions with multi-orientation and multi-object. On the other hand, 2D space also lacks the ability to capture the spatial complexity of real-world environments, especially when occlusion appears.


With the increasing availability of 3D data, research has progressively shifted toward understanding the 3D world. 3D AffordanceNet \cite{deng20213d} introduced the first benchmark dataset for learning affordances from object point clouds and proposed an end-to-end grounding architecture. Subsequent works \cite{chu2025daffordancellm,li2024laso,yu2024seqafford} continued to explore the integration of point clouds with language queries, some leveraging LLMs. However, affordance learning in embodied AI requires strong generalization capabilities, which current 3D models often fail to achieve. To address this limitation, several studies~\cite{qu2024multimodal,gao2024learning,shao2024greatgeometryintentioncollaborativeinference,lu2024geal,chu2024iris} have employed 2D affordance learning to enhance 3D affordance understanding. This approach has been successfully applied to embodied tasks such as grasping and navigation~\cite{tang2025affordgrasp,wei2025afforddexgrasp,zhang2025moma}. While 3D point clouds provide valuable geometric information for affordance analysis, they suffer from several limitations. As illustrated in~\autoref{fig:Experiment1}, the sparsity of point clouds often results in poor representation of continuous surfaces and complex structures, leading to noticeable discrepancies compared to real-world objects. Although increasing point density can improve geometric fidelity, it significantly raises computational costs. In contrast, 3D Gaussians representations not only preserve high-fidelity geometry but also enable efficient rendering, making them a more practical solution for affordance learning.

As shown in ~\autoref{tab:DatasetComparison}, existing 3D affordance datasets are primarily based on the point cloud modality. 3DAffordanceNet\cite{deng20213d} was the first large-scale benchmark for 3D point cloud affordance learning. Datasets such as LASO\cite{li2024laso} and SeqAfford\cite{yu2024seqafford} incorporate language modalities, with LASO focusing on single-question affordance answering and SeqAfford extending this to multi-question formats. PIAD\cite{yang2023grounding}, PIADv2\cite{shao2024greatgeometryintentioncollaborativeinference}, and AGPIL~\cite{zhu2025grounding3dobjectaffordance} additionally include image modalities. The PIAD family emphasizes the transfer of knowledge from 2D images to 3D affordance reasoning, while AGPIL conbined image and language together. Existing 3DGS datasets, such as CLIP-GS\cite{jiao2024clip} and ShapeSplat\cite{ma2024shapesplat}, lack affordance annotations. In contrast, our proposed 3DAffordSplat dataset is the first large-scale, multi-modal 3DGS-based affordance reasoning benchmark, incorporating point cloud, textual, and Gaussian modalities.

\subsection{Text-3DGS Cross-Modal Learning}
Text-3DGS cross-modal learning explores how textual information guide the segmentation and manipulation of 3DGS~\cite{kerbl20233d} objects. Current 3DGS semantic frameworks focuses on cross-modal feature embedding (e.g., 2D-3D, language-to-3D, etc)~\cite{qiu2024feature,ye2024gaussian,qin2024langsplat}, open-vocabulary segmentation~\cite{cen2023segment,hu2024sagd,choi2024click}, and dynamic tracking~\cite{luiten2024dynamic,shorinwa2024splatmover}.

A dominant approach is embedding 2D segmentation features into 3DGS representations to guide segmentation. Methods~\cite{ye2024gaussian,choi2024click,qin2024langsplat,zhou2024feature} projected 2D segmentation masks (from SAM~\cite{kirillov2023segment} or CLIP~\cite{radford2021learning}) into 3DGS space, leveraging them as supervision signals for object-level or part-level segmentation. These frameworks bridge the 2D-3D gap by distilling semantic priors from foundation models into spatially embedded Gaussian distributions. Gradient-Driven~\cite{joseph2024gradient} extended 2D segmentation to 3D Gaussians splats by optimizing 2D masks through gradient backpropagation and exploring affordance migration. However, it relies on precise 2D masks and selected viewpoints.

To enhance segmentation fidelity, recent works~\cite{DBLP:conf/corl/JiQZW24, qin2024langsplat, shorinwa2024splatmover} also appended additional features to Gaussians primitives and jointly optimized with those primitives parameters. These features are primarily semantic or task-specific attributes, and temporal features~\cite{li20254d} have also been explored recently.
Moreover, methods like GS-Net~\cite{zhang2024gs,ma2024shapesplat}, inspired by point cloud processing techniques, directly used Gaussian attributes as input features. This approach bypasses 2D supervision, relying instead on the inherent geometric and appearance cues of the Gaussian representation.

Unlike existing methods that embed semantic features into Gaussian primitives via parametric expansion, our AffordSplatNet dynamically generates task-specific descriptors. This enables each Gaussian primitive to adaptively respond to multiple affordance semantics based on contextual queries. This  architecture effectively addresses the challenge of multi-attribute representation, where individual Gaussian may participate in diverse affordance contexts, thereby overcoming the single-semantic limitation of conventional feature-embedding approaches.


\section{3DAffordSplat Dataset}
\label{sec:dataset}
To support our task, we introduce the first large-scale, multi-modal 3D Gaussian Splatting dataset with affordance annotations, \textbf{3DAffordSplat}, addressing the critical gap in affordance reasoning for 3DGS-based representations. Unlike existing point cloud datasets limited by sparse geometric sampling and coordinate sensitivity, our dataset leverages 3D Gaussian Splatting's inherent advantages: high-fidelity continuous surface representation (23,677 Gaussian instances) preserves fine-grained affordance details, while cross-modal alignment with 8,354 point clouds enables robust geometric reasoning. As shown in \autoref{tab:DatasetComparison}, \textbf{3DAffordSplat} uniquely provides 6,631 manually annotated affordance labels across 21 categories and 18 interaction types, paired with 15 language-guided Q\&A templates per object-affordance pair. 


\subsection{Dataset Collection}
Our 3DAffordSplat includes three modalities: 3DGS with annotations, point clouds with annotations, and language instructions. 


\noindent{\textbf{3D Gaussians}}. The 3DGS objects are sourced from ShapeSplat~\cite{ma2024shapesplat}, covering 21 categories. These Gaussians are combined with the corresponding point clouds to form 3DAffordSplat. We manually annotated part of the Gaussian data with affordances, following the standards of 3D AffordanceNet ~\cite{deng20213d}.


\noindent{\textbf{Point clouds \& Instructions.}} Our dataset builds upon the point cloud and textual data provided by~\cite{li2024laso}, selecting 21 object categories and 18 affordance types. Each object category is associated with multiple affordances, and every object-affordance pair is supplemented with a set of corresponding textual question-answer pairs. To better align the dataset with our task, we introduce a novel answer format in the instruction data. Specifically, we insert a special token ``\(\langle \text{Aff} \rangle\)'' immediately following the word denoting the affordance in each sentence, thereby enhancing the model’s ability to identify and ground affordance semantics.


\subsection{Statistics and Setting}
3DAffordSplat comprises three modalities: textual descriptions, 3D Gaussians, and point clouds. Detailed dataset statistics are provided in ~\autoref{tab:DatasetComparison}. Specifically, it covers 8,354 point clouds objects across 21 object categories and 18 affordance types with affordance annotations, with each Object-Affordance combination paired with 15 questions and 3 answers. Based on different combinations, we collected a large amount of Gaussian data, totaling 23,677 Gaussian instances, among which we manually annotated 18 Gaussians for each combination for validation and testing, amounting to 6,631 Gaussian Affordance annotations.
Following ~\cite{li2024laso}, we provide two distinct dataset settings: \textbf{Seen} and \textbf{Unseen}:
\begin{itemize}
\item \textbf{Seen}: Default configuration, where the training and testing phases share similar distributions of object classes and affordance types.

\item \textbf{Unseen}: This configuration is specifically designed to evaluate the model's ability to generalize knowledge. The test dataset has completely different Object-Affordance combinations from the training dataset. Detailed settings can be found in Appendix B.
\end{itemize}

\subsection{Pretrain and Evaluation Protocols}
During the pretrain process, each Gaussian instance is randomly assigned multiple point clouds of the same category and a question sampled from 15 template questions, along with a fixed answer relative to the Object-Affordance as the text label. During Evaluation, we use the annotated Gaussian data to ensure accurate evaluation results and use fixed multiple questions to test the model's generalization ability.


\section{AffordSplatNet}
\noindent\textbf{Task Definition.} Given a 3D Gaussian Splatting representation $\boldsymbol{\mathcal{G}} = \{\boldsymbol{m}, \boldsymbol{s}, \boldsymbol{r}, o, \boldsymbol{c}\}$, where $\boldsymbol{m} \in \mathbb{R}^3$ denotes the Gaussian center position, $\boldsymbol{s} \in \mathbb{R}^3$ represents scale parameters, and $\textbf{r} \in \mathbb{R}^4$ indicates rotation parameters (collectively termed structural features), along with opacity $o \in \mathbb{R}$ and spherical harmonics-based color features $\boldsymbol{c}$ (jointly considered as appearance features). We posit that object affordance properties primarily emerge from local structural characteristics, thus our model exclusively processes structural features $\boldsymbol{\mathcal{G}}_{\text{struct}} = \{\boldsymbol{m, s, r}\} \in \mathbb{R}^{10}$. For a textual query $Q$, the model outputs both textual response $A$ and corresponding 3D Gaussian affordance mask $\boldsymbol{\mathcal{M}} \in \{0, 1\}^N$, where $N$ denotes the number of Gaussians.


\noindent{\textbf{Preliminary.}} Given the $j$-$th$ batch of 3D Gaussian objects $\{\boldsymbol{\mathcal{G}}_i^{N_{GS}^i}\}_{i=1}^B$ with variable point counts $N_{GS}^i$, we use adaptive batch processing: 
\begin{enumerate}
    \item Downsample to the maximum number of Gaussians $N_{\text{batchmin}}^j$ in the batch to preserve structural integrity while enabling batch training, 
    \item zero padding to the maximum number of Gaussians $N_{\text{batchmax}}^j$ in the batch  for complete mask generation. 
\end{enumerate}
To leverage cross-modal alignment, each Gaussian instance $\boldsymbol{\mathcal{G}}_i$ is paired with $K$ point clouds $\boldsymbol{\mathcal{P}}=\{\boldsymbol{\mathcal{P}}_k^{N_{PC}}\}_{k=1}^K$ of matching object-affordance types, where $N_{PC}$ indicates point cloud density. The training set $\boldsymbol{\mathcal{D}}$ contains tuples $\{Q, A, \boldsymbol{\mathcal{P}}, \boldsymbol{\mathcal{G}}_{\text{struct}}^{N_{\text{batchmin}}}, \boldsymbol{\mathcal{G}}_{\text{struct}}^{N_{\text{batchmax}}}\}$.


\begin{figure*}[htbp]
    \centering  
    \includegraphics[width=1\textwidth]{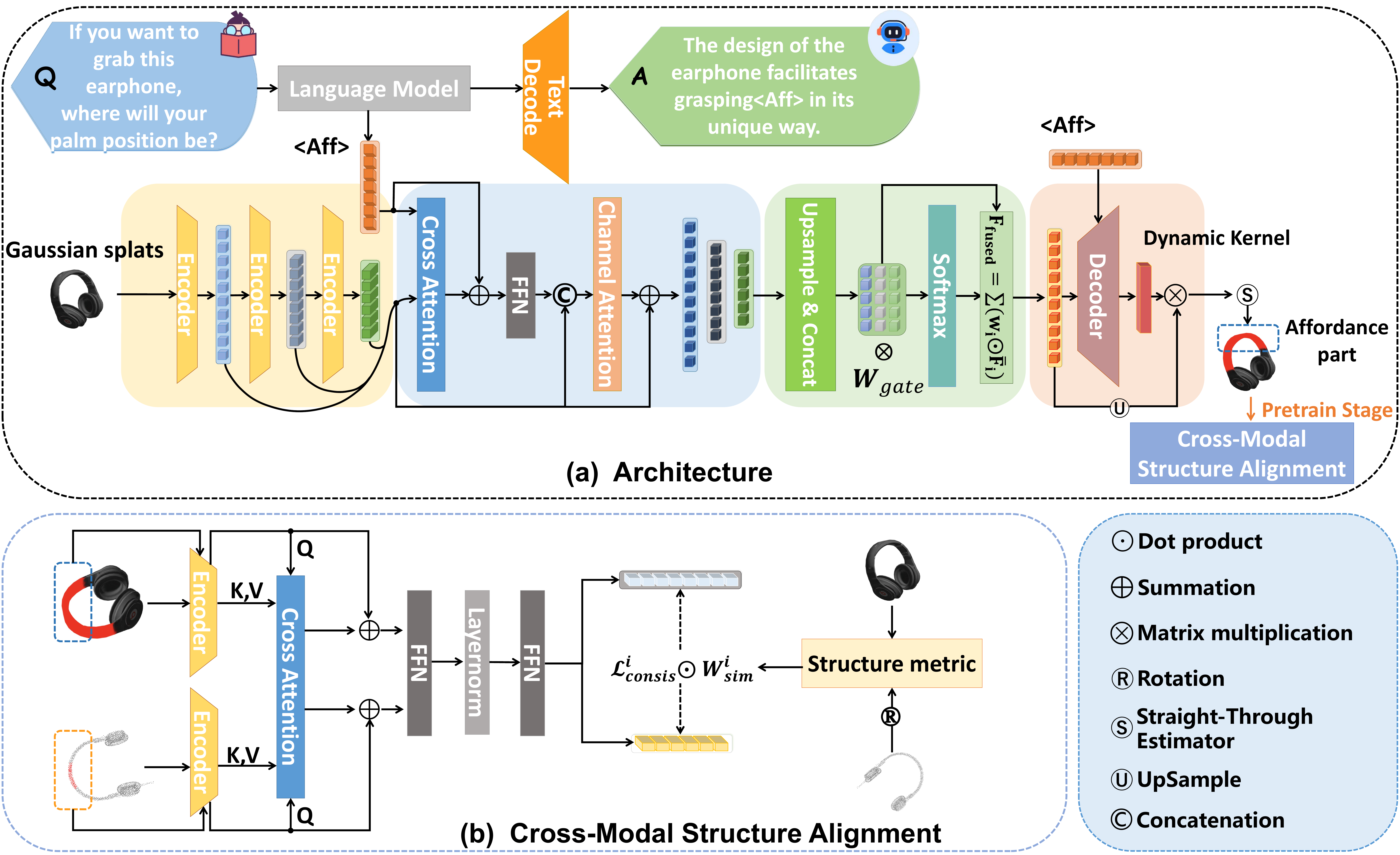}
    \caption{\textbf{Architecture Overview.} \textbf{AffordSplatNet (a)} processes 3D Gaussians and human instructions through a hierarchical pipeline. It extracts multi-granularity features from Gaussians, while a pre-trained language model infers an \(\langle \text{Aff} \rangle\) token from the text query, representing an intermediate segmentation result. These modalities are fused through attention mechanisms, with granularity selection prioritizing task-relevant spatial scales. The selected features decode into dynamic kernels for efficient affordance mask generation. To enhance 3D structural learning, \textbf{Cross-Modal Structure Alignment (CMSA) (b)} module aligns the Affordance regions and overall structural relations between the Gaussian and point cloud data at the structural level.}
    \label{fig:Architecture Overview}
\end{figure*}


\noindent{\textbf{Architecture Overview.}} The overall framework is illustrated in Fig.\ref{fig:Architecture Overview}. Given a 3D Gaussian splatting $\boldsymbol{\mathcal{G}}_{\text{struct}}$, AffordSplat utilizes  PointNet++~\cite{DBLP:journals/corr/abs-2312-00860} as 3D backbone to encode the 3D Gaussian into multi-granularity features. For a text query $Q$, a pre-trained language model (e.g., RoBERTa~\cite{DBLP:conf/nips/QiYSG17}) infers an \(\langle \text{Aff} \rangle\) token, capturing the intermediate segmentation representation from the query.  Cross-attention and channel-attention~\cite{DBLP:conf/cvpr/WangWZLZH20} mechanisms are then employed to integrates the \(\langle \text{Aff} \rangle\) token's last-layer embedding features with the Gaussian features at different granularities. The fused features are adaptively weighted through learnable granularity weights $\boldsymbol{W}_{gate}$ to dynamically select the optimal granularity. Finally, the decoder-derived dynamic kernels are convolved with the upsampled Gaussian-encoded features to produce the final Affordance mask. 

Our training process consists of two stages: Pretrain and Finetune. On the Pretrain stage, aiming to utilize a large amount of point cloud data to assist the model in learning 3DGS Affordance, we introduce a Cross-Modal Structure Alignment module to leverage large-scale point cloud affordance data. This module performs unsupervised learning by aligning the structural relations between the predicted masks and the original Gaussian models with those of the point cloud affordance regions and their corresponding point cloud models. On the Finetune stage, we employ Gaussian Affordance annotations from the 3DAffordSplat dataset for supervised training to further refine the model's performance.

\subsection{Gaussian-Text Feature Fusion}
\noindent{\textbf{Feature Encoding.}} For a given textual query $Q$, we utilize a pre-trained language model $\Psi_{LM}$ to extract the last-layer embeddings $\boldsymbol{h}_{Aff}$ of \(\langle \text{Aff} \rangle\) tokens, which encapsulates the intermediate representation for both question understanding and mask generation. This feature is projected via an $MLP$ layer $\boldsymbol{H}_{Aff} = \operatorname{MLP}(\boldsymbol{h}_{Aff}) \in \mathbb{R}^{B \times 1 \times d_{text}}$ to adapt to the subsequent modules. The language model then generates a text answer $\tilde{y}_{text}$. 

For the 3D Gaussian structural feature $\boldsymbol{\mathcal{G}}_{\text{struct}}^{N_{\text{batchmin}}}$, a hierarchical 3D encoder $\Phi_{3D}$ extracts multi-granular geometric features $\{\boldsymbol{F_{g}^{i}}\}^3_{i=1} \in \mathbb{R}^{B \times N_{i} \times d}$,  where $N_{i}$ denotes the downsampled Guassian count after the $i$-$th$ encoder stage and $d$ represents features dimension. We use the point-level feature map from the last decoding stage as the 3D backbone's output and add a transformer encoder module after the 3D encoder~\cite{DBLP:conf/nips/QiYSG17} structure for enhanced feature extraction.


\noindent{\textbf{Multi-Modal Fusion.}} We integrate linguistic features $\boldsymbol{H}_{Aff}$ and multi-granular geometric features $\{\boldsymbol{F_{g}^{i}}\}^3_{i=1}$ through cross-attention and channel-attention~\cite{DBLP:conf/cvpr/WangWZLZH20} mechanisms at spatial and channel levels. Concretely, we use $\boldsymbol{H}_{Aff}$ as queries while $\{\boldsymbol{F_{g}^{i}}\}^3_{i=1}$ as keys/values:
\begin{equation}
\boldsymbol{F}_{\text{spatial}}^{i} = \text{CrossAtt}(\boldsymbol{H}_{Aff}, \boldsymbol{F}_{g}^{i}, \boldsymbol{F}_{g}^{i}) + \text{PosEmb}(N_{i}),
\end{equation}
where $CrossAtt$ denotes cross-attention mechanism, $PosEmb$ injects position-aware cues and $\boldsymbol{F}_{\text{spatial}}^{i}\in \mathbb{R}^{B \times1 \times d_{text}}$. To enhance the discriminative power of cross-modal features, $\boldsymbol{F}_{\text{spatial}}^{i}$ is processed into $\boldsymbol{\overline{F}}_{\text{spatial}}^{i}\in \mathbb{R}^{B \times1 \times d}$  through residual connection combined with a feed-forward network (FFN). Subsequently, a channel-attention mechanism~\cite{DBLP:conf/cvpr/WangWZLZH20} adaptively recalibrates cross-modal features by fusing global linguistic context with local geometric details:
\begin{equation}
    \boldsymbol{F}_{\text{channel}}^{i} = \text{ChannelAtt}([\boldsymbol{\overline{F}}_{\text{spatial}}^{i},\boldsymbol{F_{g}^{i}}])+{\boldsymbol{F_{g}^{i}}},
\end{equation}
where $ChannelAtt$ denotes channel-attention mechanism~\cite{DBLP:conf/cvpr/WangWZLZH20} and $[\cdot]$ denotes concatenation along the channel axis, enabling joint modeling of cross-modal interactions and preserving original geometric fidelity via residual connections. 

\subsection{Granularity-Adaptive Selection and Decoder}
\noindent{\textbf{Granularity-Adaptive Selection.}} Inspired by~\cite{lu2024geal}, we integrate features across various granularities. To harmonize multi-granular geometric features, we upsample all features to a unified resolution $N$ via inverse distance weighted ($IDW$)~\cite{DBLP:conf/nips/QiYSG17} interpolation:
\begin{equation}
    \overline{\boldsymbol{F}}_i = \operatorname{IDW} (\boldsymbol{F}_{\text{channel}}^{i}) . 
    \label{eq:placeholder}
\end{equation}
Adaptive granularity selection is then achieved through learnable gating weights $\boldsymbol{W}_{gate}$:
\begin{equation}
    \boldsymbol{W} = \operatorname{Softmax} (\boldsymbol{W}_{gate}\odot\left[{\boldsymbol{\overline{F}}}_1 \| {\boldsymbol{\overline{F}}}_2 \|{\boldsymbol{\overline{F}}}_3\right]), 
\end{equation}
where $\boldsymbol{W}=\{w_i\}_{i=1}^3  \in \mathbb{R}^{B \times 3 \times d}$ satisfies $\sum_{i=1}^3{w_i^j}=1$ for each channel $j$, $\|$ denotes concatenation along the granularity axis and $\odot$ denotes element-wise multiplication, enforcing competitive allocation of importance across granularities. Final fused features combine multi-granular contributions:
\begin{equation}
    \boldsymbol{F}_{\text{fused}} = \sum_{i=1}^{3} w_i \odot \overline{\boldsymbol{F}}_{i} ,
    \label{placeholder}
\end{equation}
where $\boldsymbol{F}_{\text{fused}}\in \mathbb{R}^{B \times N \times d}$.



\noindent{\textbf{Decoder.}} The decoder module generates Gaussian-accurate affordance masks through dynamic kernel convolution and adaptive feature upsampling. First, fused multi-modal features $\boldsymbol{F}_{\text{fused}}$ are upsampled to the original Gaussian density via $IDW$ ~\cite{DBLP:conf/nips/QiYSG17}:
\begin{equation}
    \boldsymbol{F}_{up} = \operatorname{IDW} (\boldsymbol{F}_{\text{fused}}) . 
\end{equation}
where $\boldsymbol{F}_{up} \in \mathbb{R}^{B \times N_{\text{batchmax}} \times d}$.
We subsequently apply a validity mask $\boldsymbol{M}_{\text{valid}} \in \{0, 1\}^{B \times N_{\text{batchmax}}}$ to filter invalid positions:
\begin{equation}
\resizebox{0.9\linewidth}{!}{$
\boldsymbol{F}_{\text{valid}} = \boldsymbol{F}_{\text{up}} \odot \boldsymbol{M}_{\text{valid}} \quad \text{where} \quad \boldsymbol{M}_{\text{valid}}[i, j] = 
\begin{cases} 
1 & \text{if } \boldsymbol{X}_{\text{max}}[i, j] \neq 0 \\
0 & \text{otherwise}
\end{cases},
$}
\end{equation}
where $\boldsymbol{X}_{\text{max}}$ denotes positions from $\boldsymbol{\mathcal{G}}_{\text{struct}}^{N_{\text{batchmax}}}$. A transformer-based decoder then synthesizes position-aware dynamic kernels conditioned on linguistic embeddings:
\begin{equation}
    \boldsymbol{K}_{dynamic} = \operatorname{Transformer Decoder}(\boldsymbol{F}_{\text{valid}},\boldsymbol{H}_{\text{Aff}}),
\end{equation}
The final affordance mask is computed via convolution between upsampled features and dynamic kernels:
\begin{equation}
\mathcal{M}_{gs} = \sigma (\boldsymbol{F}_{\text{valid}} \ast \boldsymbol{K}_{dynamic})  \odot \boldsymbol{M}_{\text{valid}} ,
\end{equation}
where $\sigma(\cdot)$ denotes $Sigmoid$ function and $\ast$ denotes convolution.

\subsection{Cross-Modal Structure Alignment} At the pretrain stage, to leverage labeled point cloud affordance data, we propose a cross-modal structure alignment module based on structural consistency priors. For an object category, while its explicit 3D representations differ, the relative spatial relations between affordance regions and the overall structure remain invariant. 
\par
To achieve cross-modal structural alignment, we encode both the point cloud affordance regions and the Gaussian affordance regions along with their corresponding complete models into a shared  $d_{consis}$-dimensional space using modality-specific encoders:
\begin{align}
    \boldsymbol{F}_{\text{gs}}^{Aff}  = \Phi_{gs}(\overline{\mathcal{M}}_{gs} \odot\boldsymbol{\mathcal{G}}_{\text{struct}} ) ,  \boldsymbol{F}_{\text{gs}} = \Phi_{gs}( \boldsymbol{\mathcal{G}}_{\text{struct}}),
    \\
\boldsymbol{F}_{\text{pc}}^{Aff}  = \Phi_{pc}(\mathcal{M}_{pc} \odot\boldsymbol{\mathcal{P}} ) , \boldsymbol{F}_{\text{pc}} = \Phi_{pc}( \boldsymbol{\mathcal{P}}),
\end{align}    
where $\overline{\mathcal{M}}_{gs} = STE(\mathcal{M}_{gs})$, $STE$ denotes Straight-Through Estimator~\cite{liu2022nonuniform}. Then, a shared multi-head cross-attention layer computes structural affinity matrices:
\begin{equation}
    \begin{split}
        \overline{\boldsymbol{F}}_{\text{gs}} = \text{CrossAtt}(\boldsymbol{F}_{\text{gs}}^{Aff},\boldsymbol{F}_{\text{gs}},\boldsymbol{F}_{\text{gs}})
        \\
        \overline{\boldsymbol{F}}_{\text{pc}} = \text{CrossAtt}(\boldsymbol{F}_{\text{pc}}^{Aff},\boldsymbol{F}_{\text{pc}},\boldsymbol{F}_{\text{pc}}),
    \end{split}
\end{equation}
where $\boldsymbol{F}_{\text{gs}}^{Aff}$ and $\boldsymbol{F}_{\text{pc}}^{Aff}$ is used as queries, while $\boldsymbol{F}_{\text{gs}}$ and $\boldsymbol{F}_{\text{pc}}$ is used as keys/values. Affinity-aware features are projected to a latent space via shared FFNs to obtain relative structural features $\boldsymbol{Z}_{\text{gs}}$ and $\boldsymbol{Z}_{\text{pc}}$. Considering the differences in shape and structure between Gaussian objects and point cloud objects, we calculate the structural similarity between Gaussian objects and multiple point cloud objects as the weight of the loss:
\begin{equation}
    {w}_{consis}^i = \operatorname{Softmax}(-\mathcal{D}_{Chamfer}(\boldsymbol{\mathcal{G}}_{\text{struct}},\boldsymbol{\mathcal{P}}_k) / \tau),
\end{equation}
where $\mathcal{D}_{Chamfer}$ denotes Chamfer Distance~\cite{DBLP:conf/cvpr/FanSG17}  and $\tau$ is the temperature parameter.

\subsection{Training Objective}
Our framework trains a model to understand 3DGS-based affordance properties by leveraging cross-modal structural alignment during pretraining. In the pretraining phase, we focus on aligning cross-modal relative structural relations:
\begin{equation}
    \mathcal{L}_{pretrain} = \mathcal{L}_{consis},
\end{equation}
where $\mathcal{L}_{consis}$ is calculated as follows:
\begin{equation}
    \mathcal{L}_{consis} = {w}_{consis}  \odot  \mathcal{L}_{cosine},
\end{equation}
where $\mathcal{L}_{cosine}$ is the cosine loss function that aligns the relative structural relationships of affordances between Gaussian and point cloud modalities. For the fine-tuning phase, inspired by~\cite{li2024laso}, we utilize binary cross-entropy loss $\mathcal{L}_{BCE}$ and Dice loss $\mathcal{L}_{Dice}$ for affordance score prediction to addresses class imbalance and improves segmentation accuracy. Additionally, we include the text generation loss $\mathcal{L}_{text}$ for text generation: 
\begin{equation}
    \mathcal{L}_{finetune} = \mathcal{L}_{BCE} + \mathcal{L}_{Dice} + \mathcal{L}_{text},
\end{equation}
where $\mathcal{L}_{text}$ is the cross-entropy loss~\cite{liu2019roberta}.


\section{Experiments}
\subsection{Experimental Settings}
\textbf{Evaluation Metrics.} We use evaluation metrics from previous works~\cite{yang2023grounding,li2024laso,shao2024greatgeometryintentioncollaborativeinference,lu2024geal} on 3D affordance grounding to evaluate the performance on our 3DAffordSplat dataset with Seen and Unseen setting, which include Mean Intersection Over Union (mIOU)~\cite{rahman2016optimizing}, Area Under Curve (AUC)~\cite{lobo2008auc}, SIMilarity (SIM)~\cite{swain1991color} and Mean Absolute Error (MAE)~\cite{willmott2005advantages}.

\noindent{\textbf{Baseline Models.}} Since there are no works using paired point clouds-3DGS-language data to ground 3D object affordance, we select the state-of-the-art image-point clouds model, IAGNet~\cite{yang2023grounding}, and the state-of-art language-point clouds model, PointRefer~\cite{li2024laso} , as our baseline models. We evaluate them with various settings. 

\begin{table*}[t]
    \centering\small
    \setlength{\tabcolsep}{12pt}
        \caption{Evaluation on the \textbf{3DAffordSplat} dataset with various models. \texttt{FT} indicates whether fine-tuning (10 epoch) is performed when training and validation sets differ. PIADv1 and LASO are point cloud affordance datasets. $*$ is the reproduced results. }
    \begin{tabular}{c|c|cc|c|cccc} 
        \toprule
         Setting &Method &Train\&Val&Test&\texttt{FT}& mIoU↑ & AUC↑ & SIM↑ & MAE↓\\ 
        \midrule
         \multirow{12}{*}{Seen}  
         & IAGNet &PIADv1 &3DAffordSplat &$\times$ &3.64 & 48.25& 0.12& 0.22\\
         & IAGNet&PIADv1 &3DAffordSplat &\checkmark & 30.77& 79.26& 0.42& 0.20\\
         & IAGNet &PIADv1 &PIADv1 &- &21.22 &85.17 &0.56 &0.08\\
         & IAGNet &3DAffordSplat &PIADv1 &$\times$ &2.87& 51.75& 0.24& 0.20\\
         & IAGNet &3DAffordSplat &PIADv1 &\checkmark & 18.20& 84.19& 0.55& 0.09\\
         & IAGNet &3DAffordSplat &3DAffordSplat &- &31.52&81.21&0.41&0.20\\
         \cline{2-9}

         & PointRefer &LASO &3DAffordSplat &$\times$ &5.10&52.10&0.17&0.26\\
         & PointRefer &LASO &3DAffordSplat& \checkmark &49.40&93.60&0.61&0.12\\
         & PointRefer* &LASO &LASO &- &19.20&85.10&0.60&0.10\\
         & PointRefer &3DAffordSplat &LASO &$\times$ &3.80&47.40&0.19&0.23\\
         & PointRefer &3DAffordSplat &LASO & \checkmark & 18.50& 85.50& 0.60& 0.10\\
         & PointRefer &3DAffordSplat&3DAffordSplat&- &51.70&94.00&0.63&0.11\\
         
         \hline
         
         \multirow{12}{*}{UnSeen}  
         & IAGNet &PIADv1 &3DAffordSplat &$\times$ &4.19 &50.05 &0.13 &0.21\\
         & IAGNet&PIADv1 &3DAffordSplat &\checkmark &13.20&58.76&0.22&0.33\\
         & IAGNet &PIADv1 &PIADv1 &- &8.70 &73.69 &0.38 &0.11\\
         & IAGNet &3DAffordSplat &PIADv1 &$\times$ &2.16&49.42&0.24&0.20\\
         & IAGNet&3DAffordSplat &PIADv1 &\checkmark &7.38&72.46&0.35&0.12\\
         & IAGNet&3DAffordSplat&3DAffordSplat&-&7.87&53.46&0.18&0.33\\
         \cline{2-9}
         
         & PointRefer &LASO &3DAffordSplat &$\times$ &4.10&45.40&0.19&0.28\\
         & PointRefer &LASO &3DAffordSplat& \checkmark &22.20&70.40&0.34&0.28\\
         & PointRefer* &LASO &LASO &- &16.80&81.40&0.53&0.10\\
         & PointRefer &3DAffordSplat &LASO &$\times$ &3.30&47.80&0.22&0.23\\
         & PointRefer &3DAffordSplat &LASO & \checkmark &16.30& 83.50& 0.55& 0.10\\
         & PointRefer &3DAffordSplat&3DAffordSplat&-&18.30&66.50&0.28&0.28\\
        \bottomrule
    \end{tabular}
    \label{tab:Evaluation on 3DAffordSplat.}
\end{table*}


\noindent{\textbf{Implementation Details.}} 
AffordSplatNet utilizes a pretrained RoBERTa model, fine-tuned with LoRA~\cite{hu2022lora} to process language inputs. The feature dimension $d$ is set to 512. During the pretraining stage, we use unlabeled Gaussian data and labeled point cloud data for cross-modal alignment. Each Gaussian instance is randomly paired with 4 point cloud instances, generating $94,708$ Gaussian-point cloud sample pairs. We train for $1$ epoch with a learning rate of $1e-05$. On the finetune stage, We perform full fine-tuning on all components except the language module. The learning rate is set to $1e-04$, and we train for $60$ epochs. We use the AdamW optimizer at both stages to ensure stable training and effective convergence. Experiments are implemented on four GeForce RTX 4090 GPUs. 


\subsection{Evaluation on the 3DAffordSplat Dataset}
\label{sec:Evaluation on 3DAffordSplat}

We conduct comparative experiments on two baselines on different modalities to evaluate the effectiveness of the 3DAffordSplat and its cross-modal transfer performance, as shown in \autoref{tab:Evaluation on 3DAffordSplat.}.

\noindent{\textbf{High-Quality Dataset.}} Training and testing solely on point cloud datasets yields suboptimal results (e.g., mIoU of 21.22 on IAGNet-Seen and 19.20 on PointRefer-Seen), mainly due to noisy annotations in LASO~\cite{li2024laso} and PIAD~\cite{yang2023grounding} (see Appendix B: ``Dataset"). In contrast, our 3DAffordSplat dataset offers fine-grained manual labels, leading to significant performance gains after fine-tuning (e.g., mIoU of 30.77 and 49.40, respectively). The best results are achieved when both training and testing use 3DAffordSplat (e.g., mIoU of 31.52 and 51.80), underscoring the value of its high-quality annotations and well-defined setup.

\begin{figure*}[t]
    \centering  
    \includegraphics[width=1\textwidth]{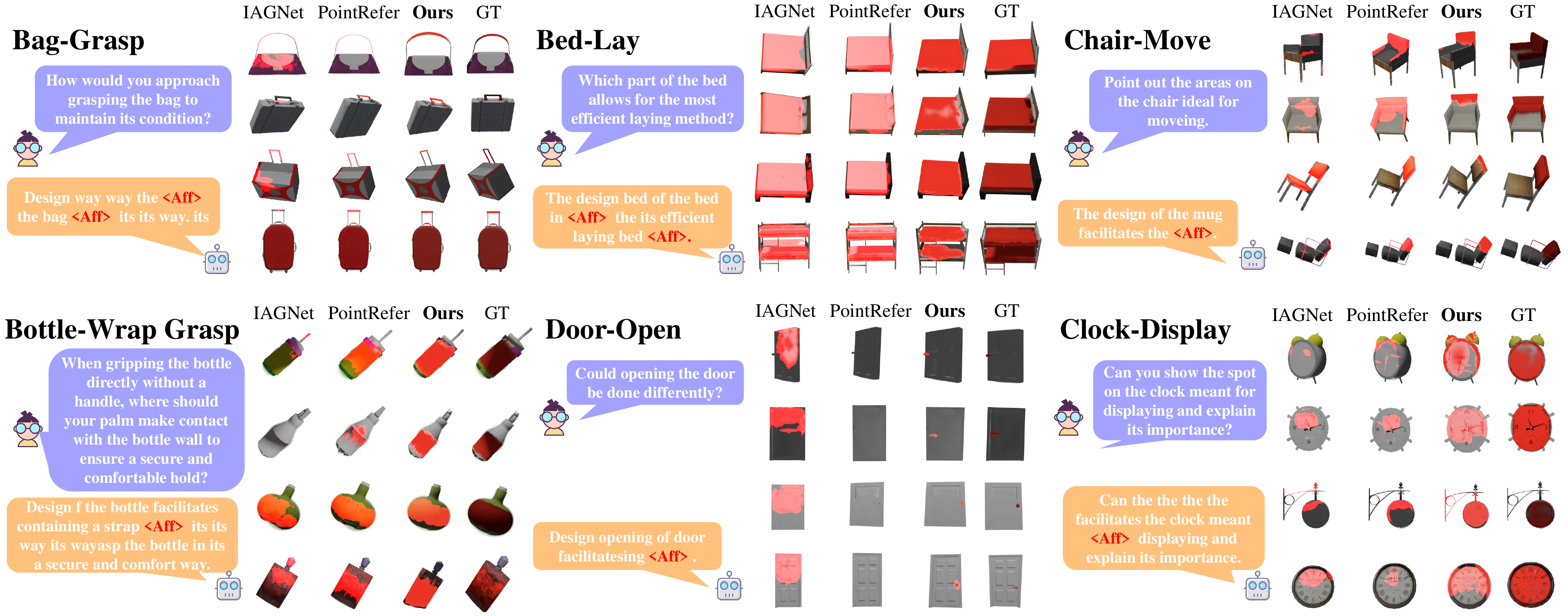}  \caption{\textbf{Visualization Results of AffordSplatNet. Each example includes one query, one answer and four object shapes, illustrating the model's generalization capability in affordance knowledge. The identified affordance regions are marked in red.}}
    \label{fig:Show-cases}
\end{figure*}

\noindent{\textbf{Efficiency in Domain Transfer}.} We conduct evaluation on cross-modality scenarios (pc→gs and gs→pc).

\noindent{\textbf{(1) pc→gs:}} Models pretrained on point clouds and fine-tuned on 3DAffordSplat show strong performance recovery, outperforming reverse modality transfer. For instance, LASO’s mIoU jumps from 5.10 to 49.40, while the reverse only improves from 3.80 to 18.50—demonstrating the superior adaptability of 3DGS.

\noindent{\textbf{(2) gs→pc:}} Compared to models trained solely on point clouds (21.22 mIoU on IAGNet-Seen, 19.20 on PointRefer-Seen), those pretrained on 3DGS and tested on point clouds achieve comparable performance (18.20 and 18.50, respectively) with reduced point cloud dependency. The 3DAffordSplat dataset boosts performance in 3DGS affordance learning while preserving the original capabilities of point cloud models.


\noindent{\textbf{Generalization Ability.}} In the UnSeen setting, all evaluation metrics are lower than in the Seen setting, highlighting the challenge of generalizing to unseen data. Although fine-tuning remains beneficial, its improvements are less substantial. For the UnSeen setting, we employ a distinct configuration, separate from those used in other datasets (see Appendix B: ``Dataset" for details). With same test set under Unseen setting, PointRefer trained on 3DAffordSplat (mIoU 7.37) achieving higher IOU, AUC, and SIM scores than those trained on LASO~\cite{li2024laso} (mIoU 4.19), demonstrating that our 3DAffordSplat dataset provides stronger support for model generalization.

\noindent{\textbf{Essential and Promising.}} Transferring from point cloud to 3DGS results in a significant performance drop without fine-tuning (e.g., PointRefer's mIoU decreases from 19.20 to 5.10 in the Seen setting), highlighting the inadequacy of point cloud knowledge for direct handling of the 3DGS modality. Fine-tuning with 3DGS significantly improves performance (e.g., PointRefer's mIoU increases from 5.10 to 49.40, and MAE drops from 0.26 to 0.12), demonstrating the necessity of 3DGS affordance datasets. Additionally, unlike the coarse, sparse annotations in point cloud datasets, 3DAffordSplat offers fine-grained, dense, and texture-rich annotations, making it promising for various downstream tasks. 

\subsection{Comparison With Baseline Models}

\noindent{\textbf{AffordSplatNet vs. Baseline Models.}} As shown in \autoref{tab:Main Result}, PointRefer~\cite{li2024laso} achieves the second-best performance across most metrics (except MAE, which overlooks structural information and cannot fully reflect affordance prediction quality) in both seen and unseen settings. This is likely due to its dual input modalities, it leverages language to infer affordance from textual instructions, enhancing task adaptability. In contrast, IAGNet~\cite{yang2023grounding} underperforms, because it emphasizes image–point cloud alignment without language guidance, limiting cross-modal generalization. It also struggles with high-dimensional Gaussian data, leading to reduced performance.

\begin{figure*}[!t]
    \centering
    \includegraphics[width=0.75\textwidth]{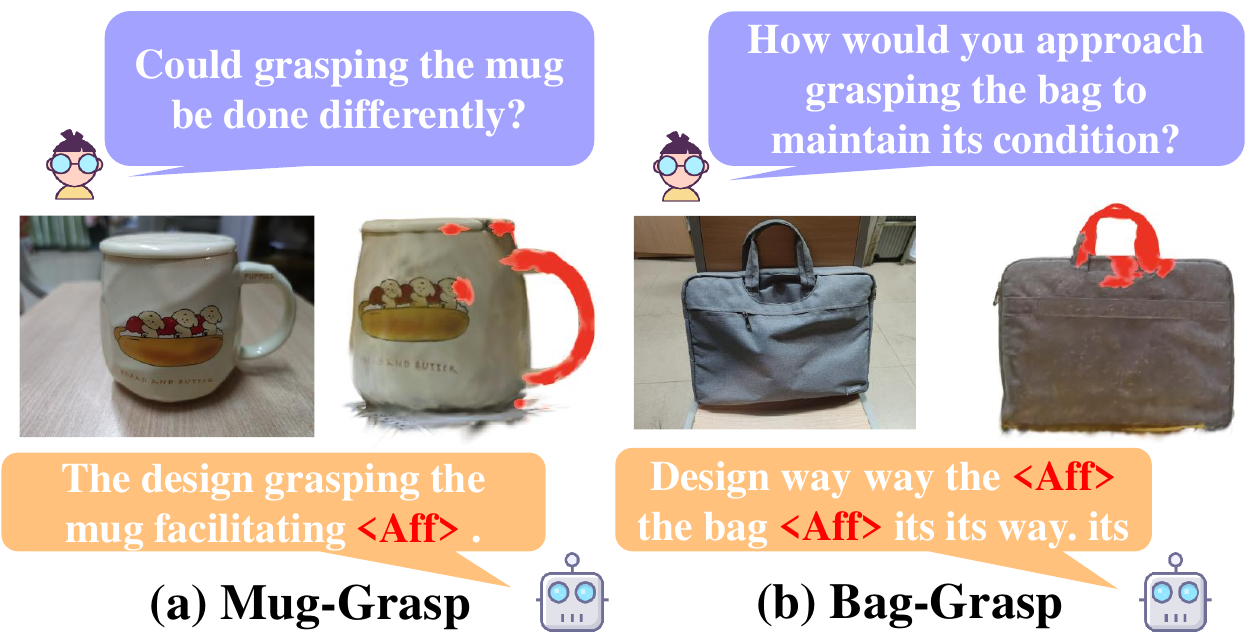}
    \caption{Real-world cases. Two common objects are shown.}
    \label{fig:Realworld}
\end{figure*}

Taking advantage of the original models' support for additional input channels, we further evaluate PointRefer and IAGNet~\cite{yang2023grounding} using \texttt{xyz} and \texttt{xyz-scale-rotate} inputs. However, a slight performance drop is observed when the scale and rotation parameters are added as extra channels. This suggests a modality gap between point cloud data and 3DGS data, indicating that models designed for point clouds may be insufficient for learning 3DGS representations.

\noindent{\textbf{Seen vs. Unseen Performance:}} All baseline models show a significant performance drop from seen to unseen settings, highlighting the challenge of generalizing affordance knowledge. In contrast, our model retains superior performance in the unseen setting, demonstrating its robustness and strong generalization capabilities, enabling it to effectively adapt to novel affordances and objects.

\begin{table}[t]
    \centering\small
    \setlength{\tabcolsep}{2pt}
\renewcommand\arraystretch{1}
        \caption{Comparison with baseline models.}
    \begin{tabular}{c|ccccc} 
        \toprule
         Setting & Method & mIoU↑ & AUC↑ & SIM↑ & MAE↓\\ 
        \midrule
         Seen &IAGNet &14.63&56.67&0.35&0.41\\
         &PointRefer & 18.40& 78.50& 0.43& 0.20\\ 
         &\textbf{AffordSplatNet (Ours)}& 30.25& 83.85& 0.44& 0.21\\
         \midrule
         Unseen &IAGNet &4.70&40.77&0.24&0.43\\ 
         &PointRefer &  15.90&  67.00&  0.31& 0.29\\ 
         &\textbf{AffordSplatNet (Ours)}& 17.31& 67.18& 0.32& 0.31\\
        \bottomrule
    \end{tabular}
    \label{tab:Main Result}
\end{table}

\subsection{Qualitative Results}

\noindent{\textbf{Case Study.}} Our model effectively interprets language instructions and accurately localizes affordance regions. By introducing a Granularity-Adaptive 3DGS architecture, it achieves robust multi-granularity affordance prediction. As illustrated in \autoref{fig:Show-cases}, 3DAffordSplatNet precisely segments fine-grained affordance components (e.g., Door-Open) while consistently capturing large continuous regions (e.g., Clock-Display). In comparison, PointRefer and IAGNet exhibit limitations such as missing regions (e.g., Door-Open), noisy predictions (e.g., Bag-Grasp), and boundary ambiguities (e.g., Bed-Lay). We attribute these shortcomings to the limited granularity adaptability of point-based representations when handling large-scale Gaussian splatting primitives.

\noindent{\textbf{Real-world Case.}} We use 3DGS~\cite{kerbl20233d} to reconstruct models in the real world with images, providing two examples with “Mug-Grasp” and “Bag-Grasp”. From \autoref{fig:Realworld}, our model can adapt to real-world objects and show promising affordance reasoning performance.  

\section{Conclusion}
In this work, we introduce 3DAffordSplat, the first large‑scale, multi‑modal affordance dataset specifically designed for 3DGS, which provides rich annotations across diverse object categories and affordance types. Based on this dataset, we propose AffordSplatNet, a novel 3DGS affordance reasoning model. By incorporating a cross‑modal structure alignment module, our model effectively bridges the gap between point‑cloud and 3DGS, yielding more accurate and robust affordance recognition. Extensive experiments demonstrate the superiority of our dataset and model, with significant improvements over existing baselines and strong generalization to unseen scenarios. In future work, we will explore integrating our affordance reasoning framework into embodied robots to physically interact with objects in dynamic environments.

\setcounter{page}{1}
\maketitlesupplementary

\begin{figure*}[htbp!]
    \centering
    \includegraphics[width=1\textwidth]{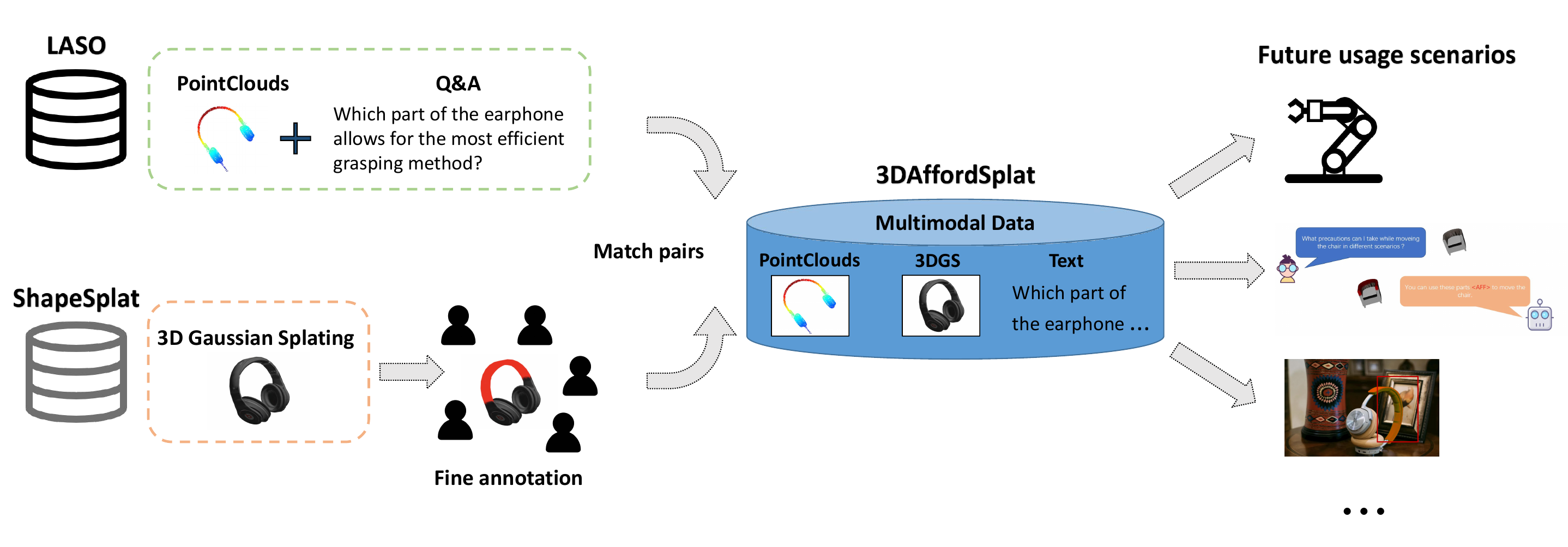}
    \caption{Dataset Construction Pipeline.}
    \label{fig:Dataset pipeline}
\end{figure*}


\section{Implementation Details}
\subsection{Method Details}

\begin{table}[htbp]\small
    \centering
    \caption{Statistics about different parameter combination.}
    \setlength{\tabcolsep}{4pt}
    \begin{tabular}{ccccc} 
        \toprule
         Method (PointRefer) & mIoU↑ & AUC↑ & SIM↑ & MAE↓\\ 
        \midrule
         xyz+rotate+scale&51.20&94.0&0.63&0.11\\
         xyz+opacity+rgb&48.40&93.8&0.61&0.11\\         xyz+rotate+scale+opacity+rgb&50.60&94.1&0.62&0.11\\
        \bottomrule
    \end{tabular}
    \label{tab:Parameter Combination}
\end{table}
In selecting the input parameters for our model, we referenced the 3D Gaussian data source from ShapeSplat~\cite{ma2024shapesplat} and conducted experiments on various parameter combinations. These experiments were performed using the PointRefer framework~\cite{li2024laso}, leveraging its \textit{add\_channel} attribute. We downsampled the 3D Gaussian data from 3DAffordSplat to 2048 points to serve as the model's input. For parameter selection, we treated the central coordinate parameters $x,y,z$ as fundamental inputs. Given the strong relationship between affordance and object structure, we categorized the remaining parameters into structural parameters (rotation and scale) and color parameters (opacity and spherical harmonics). Following ShapeSplat's approach~\cite{ma2024shapesplat}, we utilized only the first three dimensions of the color parameters, corresponding to RGB values. The experimental results, summarized in \autoref{tab:Parameter Combination}, showed that the combination of $xyz$, rotation, and scale parameters achieved the highest mIoU of 51.20. While adding opacity and RGB parameters slightly improved the AUC by 0.1, the other metrics did not perform as well. Considering the critical role of mIoU in affordance recognition, we finalized the parameter set as $xyz$, rotation, and scale for AffordSplatNet. This choice balances model performance and resource utilization effectively.

\subsection{Evaluation Metrics}
Our framework is evaluated through four key metrics that holistically assess prediction quality across spatial accuracy, distribution alignment and error magnitude:\\
\noindent{\textbf{mIoU}~\cite{rahman2016optimizing}.} The Intersection over Union (IoU) is widely recognized as the primary metric for quantifying the similarity between two shapes. It assesses how closely the predicted region aligns with the ground-truth region by calculating the ratio of their overlapping area to their combined area. The formula for IoU is expressed as:
\begin{equation}
    \text{IoU} = \frac{\text{TP}}{\text{TP} + \text{FP} + \text{FN}},
    \label{eq:iou}
\end{equation}
where $\text{TP}$ (True Positive) represents the area where the predicted region and the ground-truth region overlap, $\text{FP}$ (False Positive) indicates the area predicted but not present in the ground truth, $\text{FN}$ (False Negative) denotes the area present in the ground truth but not predicted. mIoU is the average IoU across all categories. Higher values indicate better alignment between the prediction and the ground truth.

\noindent{\textbf{AUC}~\cite{lobo2008auc}.} The Area Under the ROC Curve (AUC) is the most widely used metric for evaluating the performance of predicted saliency maps. It treats the saliency map as a binary classifier for predicting fixations across various threshold values. By measuring the true positive rate (TPR) and false positive rate (FPR) at each threshold, a ROC curve is generated. The AUC is then calculated as the integral of this curve, providing a single value that quantifies the model's ability to distinguish between positive and negative instances. Mathematically, it is expressed as:
\begin{equation}
    \text{AUC} = \int_{0}^{1} \text{TPR}(t) \, dt,
    \label{eq:auc}
\end{equation}
where $\text{TPR}(t)$ is the true positive rate at a given threshold $t$. This metric effectively summarizes the model's overall performance in predicting salient regions, with higher values indicating superior discrimination ability.

\noindent{\textbf{SIM}~\cite{swain1991color}.} The Similarity metric (SIM) evaluates the correspondence between the prediction map and the ground truth map. Given a prediction map $P$ and a continuous ground truth map $Q^D$, SIM is calculated as the cumulative sum of the minimum values at each element after normalizing the input maps:
\begin{equation}
SIM(P, Q^D) = \sum_i \min(P_i, Q^D_i),
\label{eq:similarity}
\end{equation}
where the input maps are normalized such that:
\begin{equation}
    \sum_i P_i = \sum_i Q_i^D = 1.
    \label{eq:placeholder}
\end{equation}
A higher similarity score reflects greater consistency.

\noindent{\textbf{MAE}~\cite{willmott2005advantages}.} The Mean Absolute Error (MAE) is a widely used metric in model evaluation, offering a straightforward measure of prediction accuracy. It quantifies the average magnitude of errors between the predicted and ground truth values, irrespective of their direction. Computationally, MAE aggregates the absolute differences between corresponding elements of the prediction map and the ground truth map, then normalizes this sum by the total number of elements, $n$, as expressed below:
\begin{equation}
    MAE = \frac{1}{n} \sum_{i=1}^{n} |e_i|
\label{eq:mae}
\end{equation}
Here, $e_i$ denotes the error at the i-th element, calculated as the absolute difference between the predicted and actual values. This metric penalizes larger discrepancies, lower MAE values indicate superior performance.

In summary, these metrics offers a comprehensive evaluation framework for affordance prediction models. An ideal model should achieve high mIoU, high AUC, high SIM, and low MAE.

\section{Dataset}
\subsection{3DAffordSplat}
As shown in \autoref{fig:Dataset pipeline}, our 3DAffordSplat dataset integrates data from LASO~\cite{li2024laso} and ShapeSplat~\cite{ma2024shapesplat}. The point cloud and textual data are sourced from LASO~\cite{li2024laso}, while the 3D Gaussian data is derived from ShapeSplat~\cite{ma2024shapesplat}.

\noindent{\textbf{3D Gaussians.}} Our 3D Gaussian objects are generated from a subset of ShapeSplatv1~\cite{ma2024shapesplat}. ShapeSplatv1's Gaussian data is generated from two primary sources: ModelNet~\cite{ma2024shapesplat} and ShapeNet~\cite{ma2024shapesplat}. These sources produce two sub-datasets within ShapeSplatv1~\cite{ma2024shapesplat}:
\begin{itemize}
    \item {ModelSplat~\cite{ma2024shapesplat}: Derived from ModelNet~\cite{wu20153d}, where "door" and "vase." data derived from. }
    \item {ShapeSplat~\cite{ma2024shapesplat}: Derived from ShapeNet~\cite{chang2015shapenet}, this sub-dataset covers the majority of our Gaussian objects.}
\end{itemize}
According to the standard of ~\cite{deng20213d}, we manually labeled a small part of the Gaussian datas.

\begin{figure*}[t]
    \centering  
    \includegraphics[width=1\textwidth]{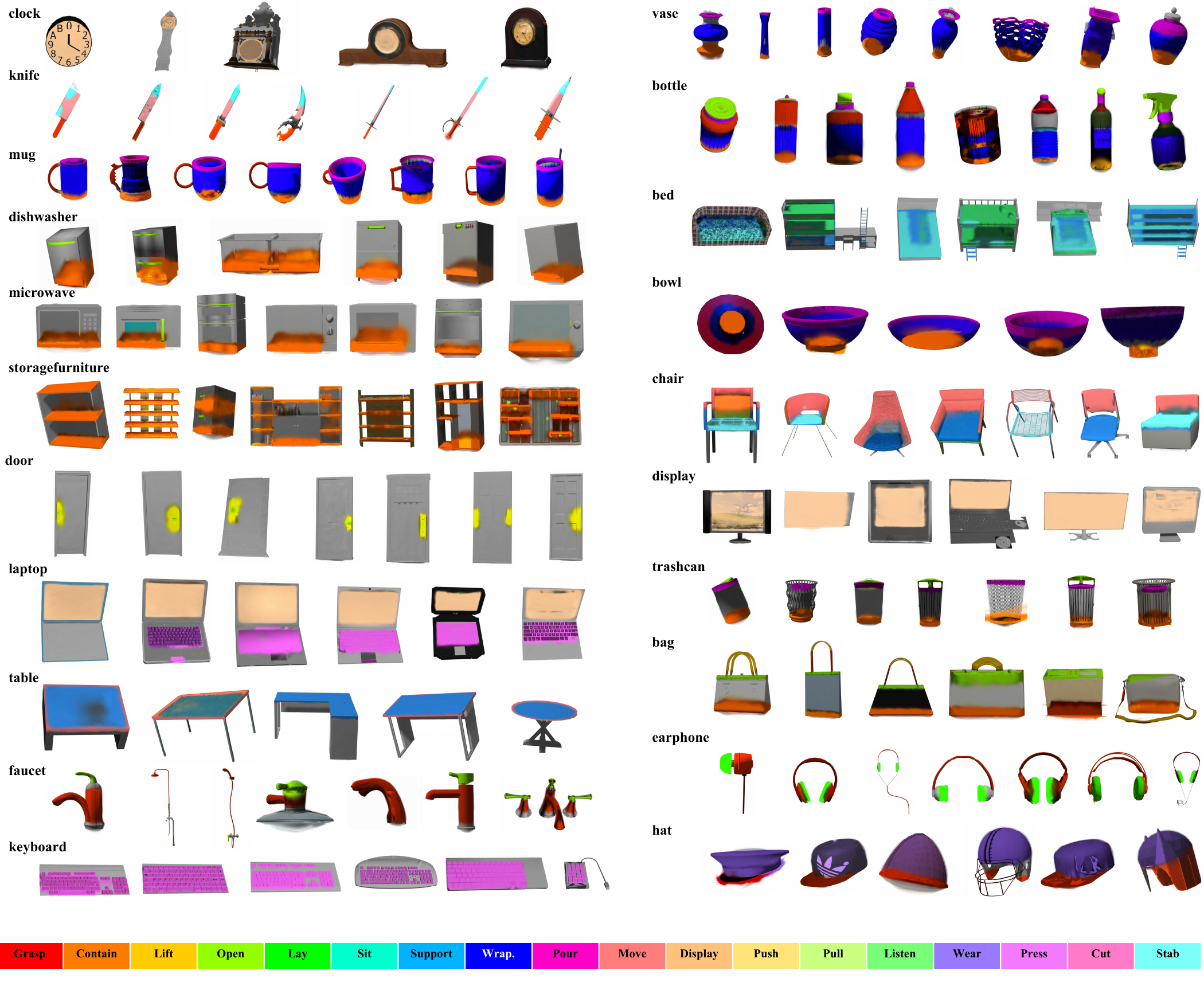}
    \caption{\textbf{Annotated Examples.}}
    \label{fig:Annotated Example}
\end{figure*}

\noindent{\textbf{Point Clouds and Text.}} Since ShapeSplat~\cite{ma2024shapesplat} lacks Gaussian objects for LASO's~\cite{li2024laso} "scissors" and "refrigerator" categories, these were excluded. After aligning the datasets, we merged them to create our multimodal 3DAffordSplat dataset. Each data instance includes three modalities: point cloud, 3D Gaussian, and text. The dataset comprises 21 object categories and 18 affordance classes, supporting applications like prediction, embodied question answering, and interactive grasping. Detailed statistics are provided in \autoref{tab:Affordance Statistics}, and annotated examples are shown in \autoref{fig:Annotated Example}.

\begin{figure}[htbp]
    \centering
    \includegraphics[width=0.48\textwidth]{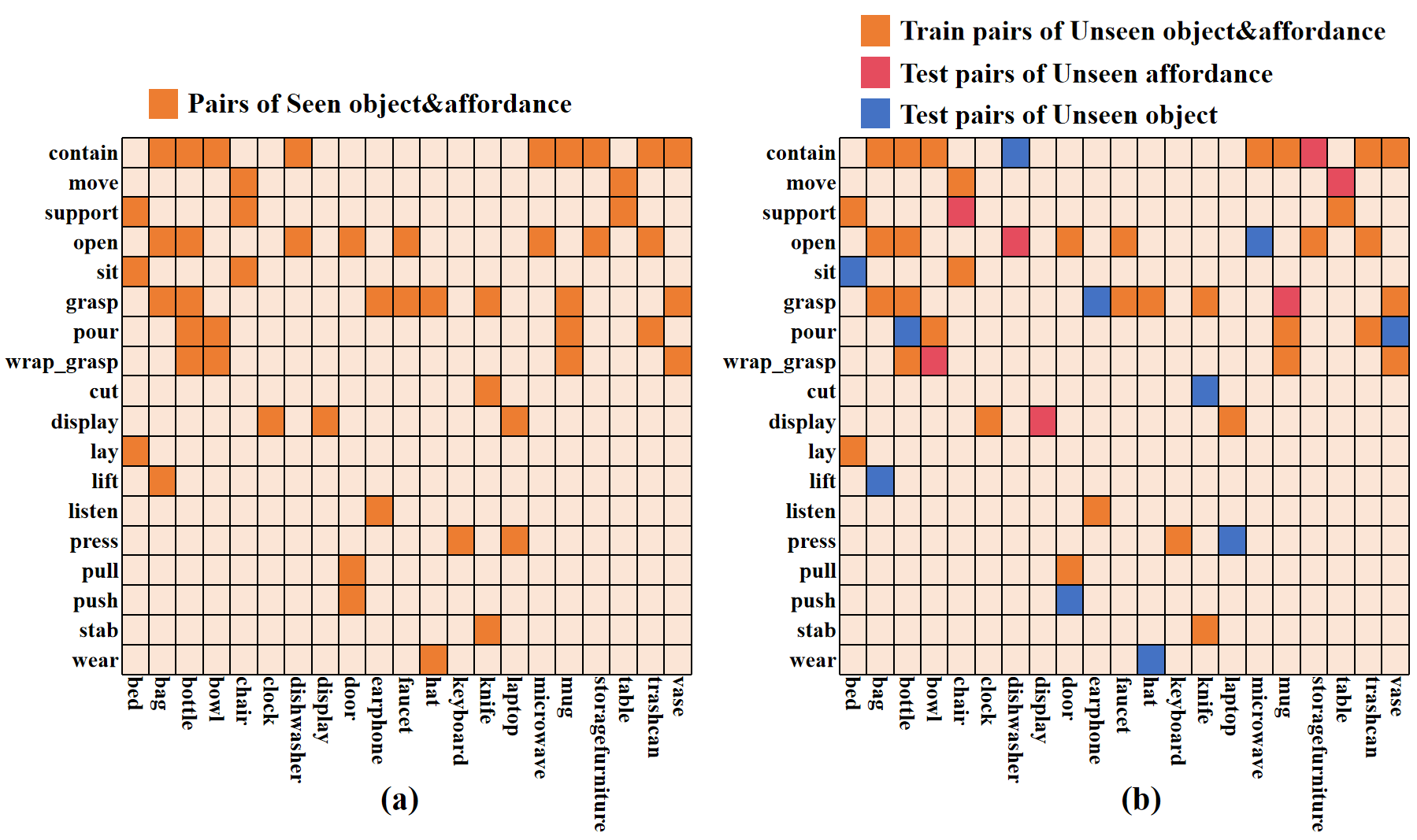}
    \caption{Seen (a) and UnSeen (b) Setting. }
    \label{fig:Seen_and_Unseen_set}
\end{figure}

\noindent{\textbf{Seen and UnSeen setting.}} \autoref{fig:Seen_and_Unseen_set} shows the details that how we design the Seen and UnSeen setting for our dataset. Our dataset's design follows the conventional approach~\cite{li2024laso,yang2023grounding, yu2024seqafford} used in most datasets, where the Seen setting ensures consistency between the training and testing data distributions and UnSeen setting aims to validate the generalization ability of the model. However, our dataset innovates by introducing a novel UnSeen configuration. For 3DAffordSplat, in the Seen setting, the training and testing sets share the same distributions of object classes and affordance types, ensuring stability in model evaluation. For the UnSeen setting, we specifically design the dataset to evaluate the model's ability to generalize to unseen object types, affordance types, and object-affordance combinations. This configuration tests how well the model can adapt to scenarios not encountered during training. For instance, object types like "Display," affordance types like "lift," and object-affordance combinations like "mug-grasp" are exclusively present in the testing and validation sets, ensuring a rigorous assessment of the model's generalization capabilities. This design highlights our dataset's focus on real-world applicability and robustness.
\begin{table*}[htbp!]\small
    \centering
        \caption{Statistics about affordance categories, 3DGS and point clouds in 3DAffordSplat.}
            \vspace{-10pt}
        \setlength{\tabcolsep}{12pt}
    \begin{tabular}{c|c|cc}
    \toprule
        \hline
        Object & Affordance & $Num_{GS}$ & $Num_{PC}$\\
    \midrule
        \hline
        Bag & grasp, lift, contain, open & 83&100\\
        Bed & lay, sit, support & 233&145\\
        Bottle & contain, open, wrap - grasp, grasp, pour & 498&328\\
        Bowl & contain, wrap - grasp, pour & 186&150\\
        Chair & sit, support, move & 6,731&1886\\
        Clock & display & 651&353\\
        Dishwasher & open, contain &93&132\\
        Display & display &1,091&488\\
        Door & open, push, pull &129&175 \\
        Earphone & grasp, listen &73&178\\
        Faucet & grasp, open &744&359 \\
        Hat & grasp, wear &218&177 \\
        Keyboard & press &65&125\\
        Knife & grasp, cut, stab &423&255\\
        Laptop & display, press &460&337\\
        Microwave & open, contain, support &152&148\\
        Mug & contain, pour, wrap - grasp, grasp &214&151\\
        Storage Furniture & contain, open &2,321&690\\
        Table & support, move &8,390&1420\\
        Trash Can & contain, pour, open & 342&251 \\
        Vase & contain, pour, wrap - grasp &575&383\\
        Total & - &23,672 (6,631)&8,231\\
        \hline
    \bottomrule
    \end{tabular}
    \label{tab:Affordance Statistics}
\end{table*}

\subsection{LASO and IAGNet}
\begin{figure}[t]
    \centering  
    \includegraphics[width=0.45\textwidth]{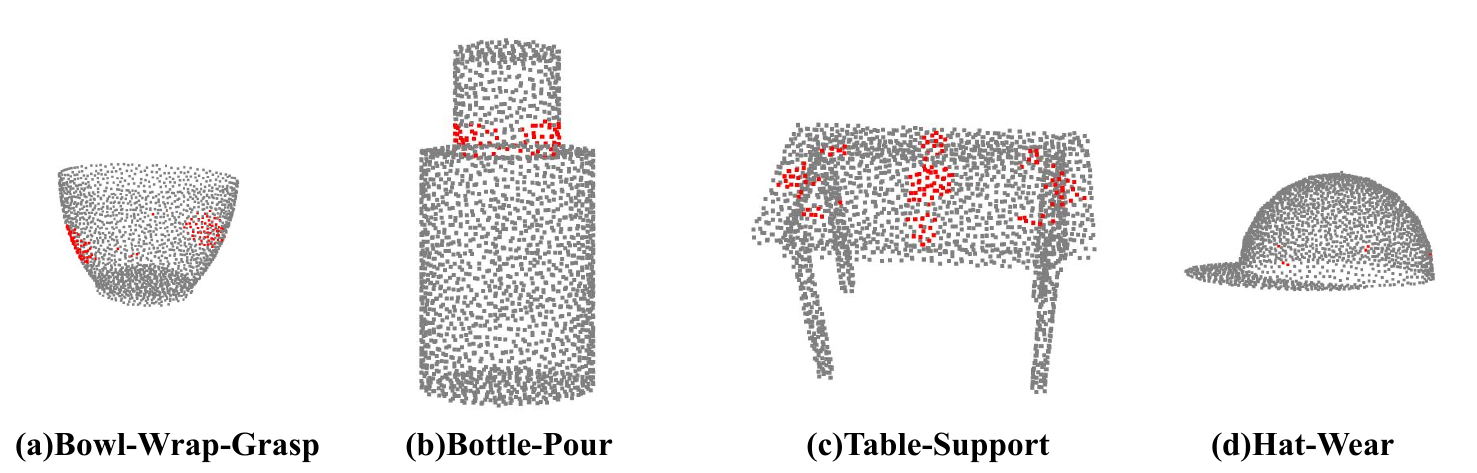}
        \vspace{-10pt}
    \caption{Examples of problematic labels in 3DAffordanceNet.}
        \vspace{-10pt}   
    \label{fig:bad-example}
\end{figure}

The point cloud data employed in both LASO~\cite{li2024laso} and IAGNet~\cite{zhu2025grounding3dobjectaffordance} originates from 3D AffordanceNet. As shown in \autoref{fig:bad-example}, during our data curation process, we found several notable issues within this dataset, including incomplete annotations (e.g., cases (a), (c), and (d)) and labeling errors (e.g., case (b)).

\section{Experiments}
\subsection{Details of Datasets validation}
\cref{sec:Evaluation on 3DAffordSplat} explores the validity of the 3DAffordSplat dataset. PointRefer~\cite{li2024laso} is a point cloud - language affordance model and IAGNet~\cite{zhu2025grounding3dobjectaffordance} is a point cloud - image affordance model. When working with the 3DAffordSplat dataset, we replace the required input point cloud modality with the 3D Gaussian data from 3DAffordSplat.

To evaluate the performance of our dataset across different models, we ensure consistency by setting the input dimensions for both PointRefer~\cite{li2024laso} and IAGNet~\cite{zhu2025grounding3dobjectaffordance} to their default 2048 points. Specifically, we sample 3D Gaussian objects from the 3DAffordSplat dataset to 2048 points before feeding them into the models.
For training, we adhere to the default settings of each model. Both PointRefer and IAGNet utilize a batch size of 16 and a learning rate of 1e-4, with the feature dimension $d$ set to 512. When fine-tuning is not required, the seen/unseen splits follow the train/test dataset's defined split. Conversely, when fine-tuning is necessary, the seen/unseen splits adhere to the validation dataset's defined split. This approach ensures a fair comparison across all datasets.

\subsection{Details of Modular Baselines}
We compare two representative open-source baselines in our experiments:
\textbf{(1) PointRefer~\cite{li2024laso}} – the current state-of-the-art model for language-to-point cloud affordance prediction, focusing on cross-modal alignment between text and 3D point cloud.
\textbf{(2) IAGNet~\cite{zhu2025grounding3dobjectaffordance}} – a strong model designed for image-to-point cloud affordance learning.

We follow the baselines' original implementation settings and replace their point cloud modality with our Gaussian-based representation for training and evaluation.  Both of them are trained on 3DAffordSplat with the same epoch of our own model, following the Seen/Unseen setting of 3DAffordSplat. As for training details, we follow the default settiing of their own. Both PointRefer and IAGNet have their batch-size set to 16, with a learning rate of 1e-4. The feature dimension \textit{d} is set to 512.

According to the experimental results: \textbf{PointRefer} shows relatively good adaptability to our task and Gaussian modality, especially when fine-tuned. However, it struggles with detecting small or fine-grained objects, and exhibits difficulty in producing continuous affordance surfaces, which are essential for more precise interaction understanding. \textbf{IAGNet}, while effective on standard point clouds, performs poorly on our Gaussian modality, particularly when the number of sample points and the input dimension increases. This is mainly because this model rely on the pair image heavily, lacking the architectural flexibility to handle densely, complex surface of 3DGS.

\subsection{Metrics of Each Object and Affordance}
As shown in \autoref{tab:Affordance Evaluation Statistics} and \autoref{tab:Object Evaluation Statistics}, we provide detailed metric results for our AffordSplatNet model, listed separately by the categories of object and affordance. 


\begin{table}[htbp]
  \centering
  \caption{Affordance Evaluation Statistics}
        \vspace{-10pt}
      \small
    \setlength{\tabcolsep}{2pt}
  \begin{tabular}{c|cccccc}
     \toprule
  Affordance & {MAE$\downarrow$} & {SIM$\uparrow$} & {KLD$\downarrow$} & {AUC$\uparrow$} & {IOU$\uparrow$} & {Num} \\
      \midrule
    pour        & 0.1509 & 0.5026 & 1.0117 & 0.9489 & 0.4239 & 65.0 \\
    contain     & 0.1827 & 0.4452 & 1.4654 & 0.8368 & 0.3257 & 107.0 \\
    open        & 0.1212 & 0.2733 & 2.3809 & 0.8359 & 0.1696 & 71.0 \\
    display     & 0.2579 & 0.2748 & 1.9491 & 0.7809 & 0.1443 & 39.0 \\
    press       & 0.2944 & 0.2922 & 2.2169 & 0.6214 & 0.1625 & 17.0 \\
    lay         & 0.2306 & 0.4138 & 1.6142 & 0.7639 & 0.1858 & 13.0 \\
    support     & 0.2182 & 0.5505 & 0.9475 & 0.8723 & 0.3786 & 59.0 \\
    sit         & 0.2041 & 0.3915 & 1.4341 & 0.8418 & 0.2710 & 35.0 \\
    stab        & 0.1159 & 0.5757 & 0.6676 & 0.9806 & 0.4535 & 13.0 \\
    grasp       & 0.2294 & 0.4753 & 1.1450 & 0.8388 & 0.3514 & 80.0 \\
    cut         & 0.2387 & 0.6888 & 0.5459 & 0.9962 & 0.4852 & 13.0 \\
    wrap\_grasp & 0.2842 & 0.4768 & 1.7076 & 0.7441 & 0.3427 & 52.0 \\
    move        & 0.1934 & 0.6621 & 0.5422 & 0.9460 & 0.5395 & 46.0 \\
    lift        & 0.1098 & 0.3888 & 1.4337 & 0.9519 & 0.3749 & 8.0 \\
    listen      & 0.4539 & 0.3604 & 1.2148 & 0.4640 & 0.1251 & 7.0 \\
    wear        & 0.2827 & 0.8322 & 0.1999 & 0.9666 & 0.6370 & 13.0 \\
    pull        & 0.0083 & 0.4952 & 1.4048 & 0.9612 & 0.3607 & 9.0 \\
    push        & 0.0619 & 0.2522 & 3.5051 & 0.6516 & 0.1354 & 9.0 \\
      \bottomrule
    \end{tabular}
    \label{tab:Affordance Evaluation Statistics}
\end{table}


\noindent{\textbf{Affordance Evaluation Statistics.}}Affordances with clear spatial structures, such as cut, wear, stab, pour and pull, achieve excellent scores across all metrics, with low MAE (e.g., stab: 0.1159, pull: 0.0083), high SIM (e.g., wear: 0.8322), and high IOU values. This shows that our model is good at dealing with affordance with typical structure. Affordances involving interactions like move, grasp and lift, also get strong results, indicating the dataset's capacity to represent fine-grained spatial-functional patterns. More ambiguous affordances, such as press, listen, push, and display, show relatively lower scores, which may reflect the complexity or variability of these interactions across objects.


\begin{table}[htbp]
  \centering
      \caption{Object Evaluation Statistics}
          \vspace{-10pt}
      \small
    \setlength{\tabcolsep}{2pt}
  \begin{tabular}{c|cccccc}
     \toprule
  Affordance & {MAE$\downarrow$} & {SIM$\uparrow$} & {KLD$\downarrow$} & {AUC$\uparrow$} & {IOU$\uparrow$} & {Num} \\
      \midrule
    trashcan         & 0.1530 & 0.3486 & 1.8340 & 0.8560 & 0.2623 & 36.0 \\
    clock            & 0.2812 & 0.1975 & 2.6751 & 0.7662 & 0.0957 & 13.0 \\
    keyboard         & 0.3486 & 0.5971 & 0.7169 & 0.7931 & 0.3785 &  6.0 \\
    bed              & 0.2181 & 0.3346 & 1.7756 & 0.7635 & 0.1660 & 38.0 \\
    dishwasher       & 0.0973 & 0.4337 & 1.6988 & 0.9551 & 0.3102 & 17.0 \\
    knife            & 0.1931 & 0.6236 & 0.6251 & 0.9584 & 0.4563 & 39.0 \\
    bottle           & 0.1760 & 0.2242 & 2.8038 & 0.6791 & 0.1513 & 64.0 \\
    chair            & 0.1819 & 0.6192 & 0.7054 & 0.9242 & 0.4704 & 69.0 \\
    bag              & 0.1639 & 0.3263 & 1.6257 & 0.8942 & 0.2426 & 32.0 \\
    laptop           & 0.2127 & 0.2265 & 2.1474 & 0.7478 & 0.1397 & 24.0 \\
    table            & 0.2461 & 0.5778 & 0.7799 & 0.9043 & 0.4411 & 46.0 \\
    faucet           & 0.2575 & 0.5271 & 1.0606 & 0.8000 & 0.3730 & 26.0 \\
    earphone         & 0.4239 & 0.4640 & 1.1778 & 0.6258 & 0.2359 & 14.0 \\
    storagefurniture & 0.2876 & 0.3986 & 2.0061 & 0.7716 & 0.1779 & 17.0 \\
    hat              & 0.2337 & 0.6980 & 0.4254 & 0.9638 & 0.5358 & 26.0 \\
    microwave        & 0.1536 & 0.3076 & 2.3141 & 0.8031 & 0.1941 & 19.0 \\
    mug              & 0.1522 & 0.5574 & 0.7010 & 0.9605 & 0.4517 & 52.0 \\
    bowl             & 0.2908 & 0.4388 & 1.4375 & 0.7582 & 0.3451 & 39.0 \\
    vase             & 0.1452 & 0.6389 & 0.5255 & 0.9697 & 0.5416 & 39.0 \\
    door             & 0.0263 & 0.4116 & 2.1326 & 0.8581 & 0.2791 & 27.0 \\
    display          & 0.3256 & 0.3152 & 1.7760 & 0.6424 & 0.1171 & 13.0 \\
      \bottomrule
    \end{tabular}
    \label{tab:Object Evaluation Statistics}
\end{table}
\noindent{\textbf{Object-Level Evaluation.}}Objects with clear, typical geometries such as knife, hat, chair, vase, and door achieve consistently strong performance. For example, hat reaches an IOU of 0.5358 and a SIM of 0.6980, while door yields the lowest MAE (0.0263).
Objects supporting multiple affordances such as table, microwave, and faucet, also demonstrate robust scores. In contrast, classes with fewer samples or higher shape variation (e.g., clock) see relatively lower performance, suggesting opportunities for future dataset expansion or balancing.

\subsection{More Experiments}
To evaluate the contributions of individual component within our model, we conduct an ablation study on two key modules: the language module and the alignment module. The ablation results are shown in \autoref{tab:Language Ablation} and \autoref{tab:Ablative Results}.  

\begin{table}[t]
    \label{tab:Language Ablation}
    \centering
    \small
    \setlength{\tabcolsep}{5pt}
          \caption{Ablation study on various language encoders.}
                  \vspace{-10pt}
    \begin{tabular}{c|cccc}
    \toprule
         Language Encoder&  \textit{mIOU}$\uparrow$&  \textit{AUC}$\uparrow$& \textit{SIM}$\uparrow$& \textit{MAE}$\downarrow$\\
    \hline
            Bart (Decoder-only)& 20.61& 73.52& 0.35& 0.27\\
            GPT2 (Encoder-Decoder)& 32.96& 81.34& 0.44& 0.22\\
            Roberta (Encoder-only)& 33.03& 84.67& 0.46& 0.21\\
    \bottomrule
    \end{tabular}    \vspace{-10pt}\label{tab:Language Ablation}
\end{table}


\begin{table}[t]
    \label{tab:Ablative Results}
    \centering
    \small
    \setlength{\tabcolsep}{1pt}
         \caption{Ablation results on the 3DAffordSplat dataset.}
                 \vspace{-10pt}
    \begin{tabular}{c|cc|cccc}
    \toprule
         Setting&  Variants&  Task&  \textit{mIOU}$\uparrow$&  \textit{AUC}$\uparrow$& \textit{SIM}$\uparrow$& \textit{MAE}$\downarrow$\\
    \midrule
    \multirow{2}{*}{Seen}& Ours     & Pretrain-finetune    & 33.03& 84.67& 0.46& 0.21\\   
                         & w/o CMSA & Finetune             & 37.18& 81.34& 0.48& 0.20\\
    \midrule
    \multirow{2}{*}{UnSeen}& Ours     & Pretrain-finetune    & 18.91& 66.71& 0.32& 0.31\\   
                         & w/o CMSA   & Finetune             & 17.93& 62.39& 0.27& 0.31\\
    \bottomrule
    \end{tabular}    \vspace{-10pt}\label{tab:Ablative Results}
\end{table}

\noindent{\textbf{Ablation on language encoders.}} Since our model is language-guided, we first evaluate three language backbones on our model. Specifically, we compare RoBERTa~\cite{liu2019roberta} (encoder-only), GPT-2 ~\cite{lagler2013gpt2} (decoder-only), and BART~\cite{lewis2019bart} (encoder-decoder). Among these, RoBERTa (mIoU=33.03) achieves the best overall performance, followed by GPT-2 (mIoU=32.96). The strong performance of RoBERTa may be its efficient bidirectional contextual encoding and its adaptive to Multimodal Large Language Model (MLLM), which captures task-relevant semantics effectively. GPT-2, while slightly less accurate, its generative capacity makes it suitable for instruction-conditioned task. But since it is a generative model, its answer may away from reasoning. In contrast, BART (mIoU=20.61) performs the worst in our setting and also takes the longest time to train, maybe its decoder-only structure doesn't combined with visual features well and performs less well.

\noindent{\textbf{Ablation on alignment module.}} The CMSA module demonstrates significant value in unseen object scenarios (mIOU: 18.91 $\xrightarrow{}$ 17.93, AUC: 66.71 $\xrightarrow{}$ 17.93, SIM: 0.32 $\xrightarrow{}$ 0.27 drop without CMSA). This aligns with findings in cross-modal representation learning~\cite{zhang2024pcalign,yang2023grounding, xue2023ulip}, where alignment mechanisms bridge heterogeneous feature spaces (point clouds $\leftrightarrow$ 3D Gaussians). Key factors as follows: (1) CMSA maps local geometric features to a unified semantic space, enabling transfer of affordance priors learned during pre-training (e.g., "graspable" regions on diverse objects). (2) Pre-trained alignment acts as a knowledge bottleneck, filtering task-irrelevant geometric variations while preserving affordance-critical patterns. This compensates for the absence of fine-tuning data for unseen objects.

Contrary to expectations, removing CMSA improves mIOU for seen objects (33.03 $\xrightarrow{}$ 37.18). This paradox highlights two phenomena:
\begin{itemize}
    \item \textbf{Task-Specific Overalignment}: Pre-trained alignment may enforce overly rigid feature correspondences, conflicting with fine-tuning data, which means excessive cross-modal constraints can suppress task-specific feature adaptations (e.g., prioritizing affordance metrics like SIM over raw geometric accuracy).
    \item \textbf{Data Sufficiency Mitigation}: For seen objects, abundant fine-tuning data likely overshadows pre-training benefits.
\end{itemize}

Overall, the alignment mechanism plays a crucial role in bridging point cloud features and 3D Gaussian features. Without CMSA, the model fails to acquire basic affordance knowledge from point clouds to transfer it into 3D Gaussian. 

\subsection{More Visualization Results}
\begin{figure*}[htbp!]
    \centering  
    \includegraphics[width=1\textwidth]{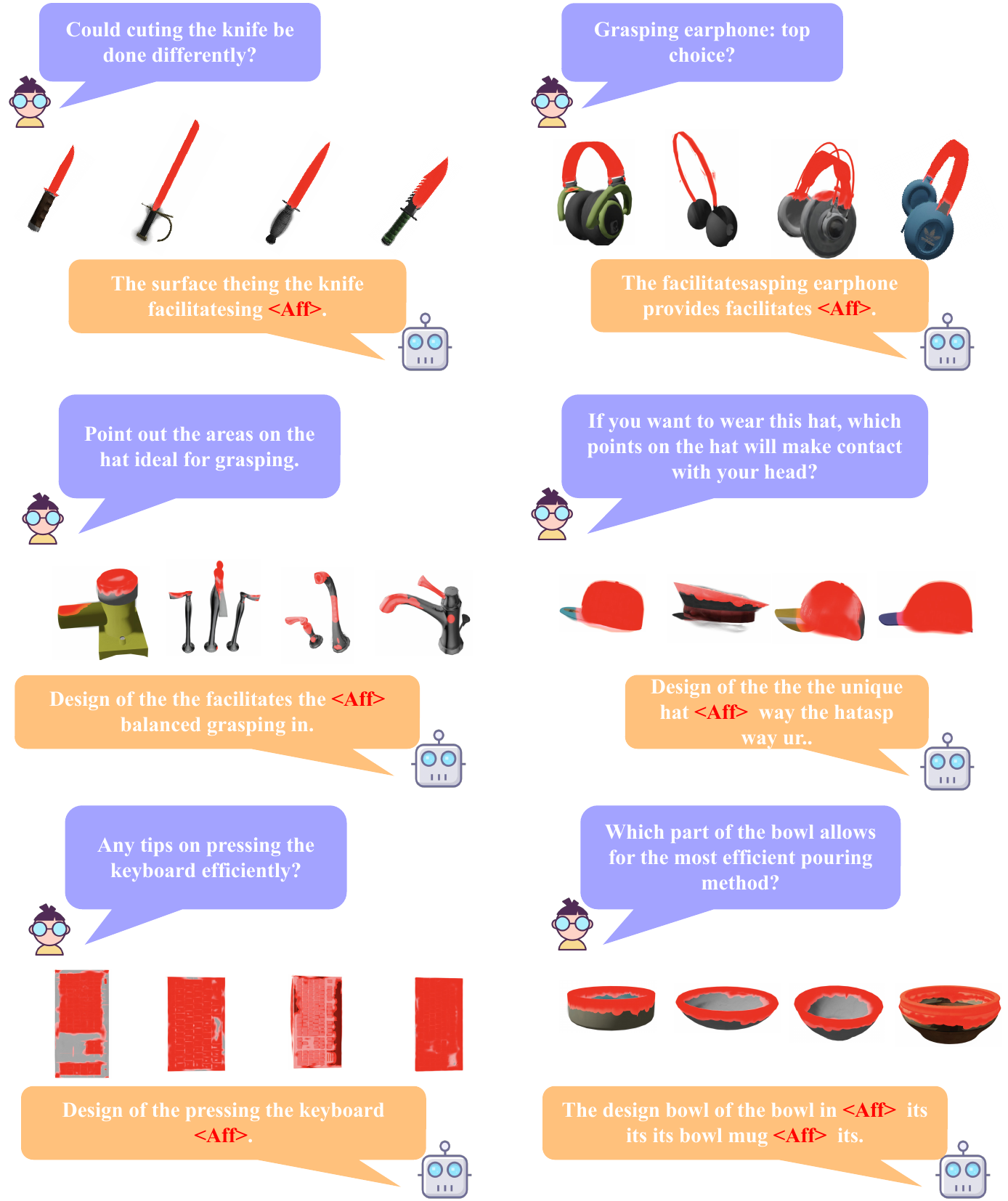}
    \caption{\textbf{Visualization Results1.}}
    \label{fig:Show-cases-col1}
\end{figure*}

\begin{figure*}[t]
    \centering  
    \includegraphics[width=1\textwidth]{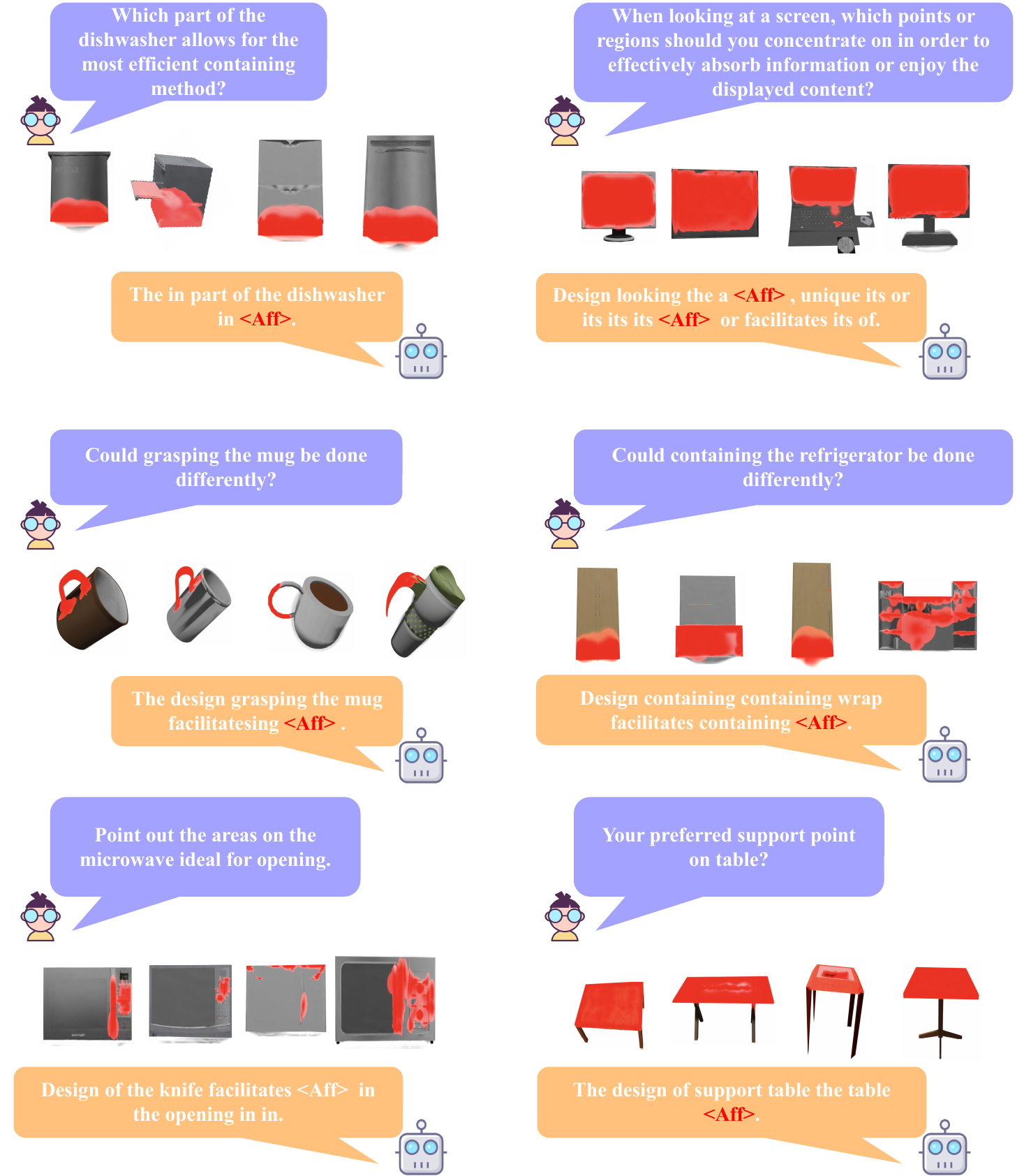}
    \caption{\textbf{Visualization Results2.}}
    \label{fig:Show-cases-col2}
\end{figure*}

More visualization results of the affordance prediction from our AffordSplatNet are shown in \autoref{fig:Show-cases-col1} and \autoref{fig:Show-cases-col2}.
\begin{figure}[t]
    \centering  
    \includegraphics[width=0.45\textwidth]{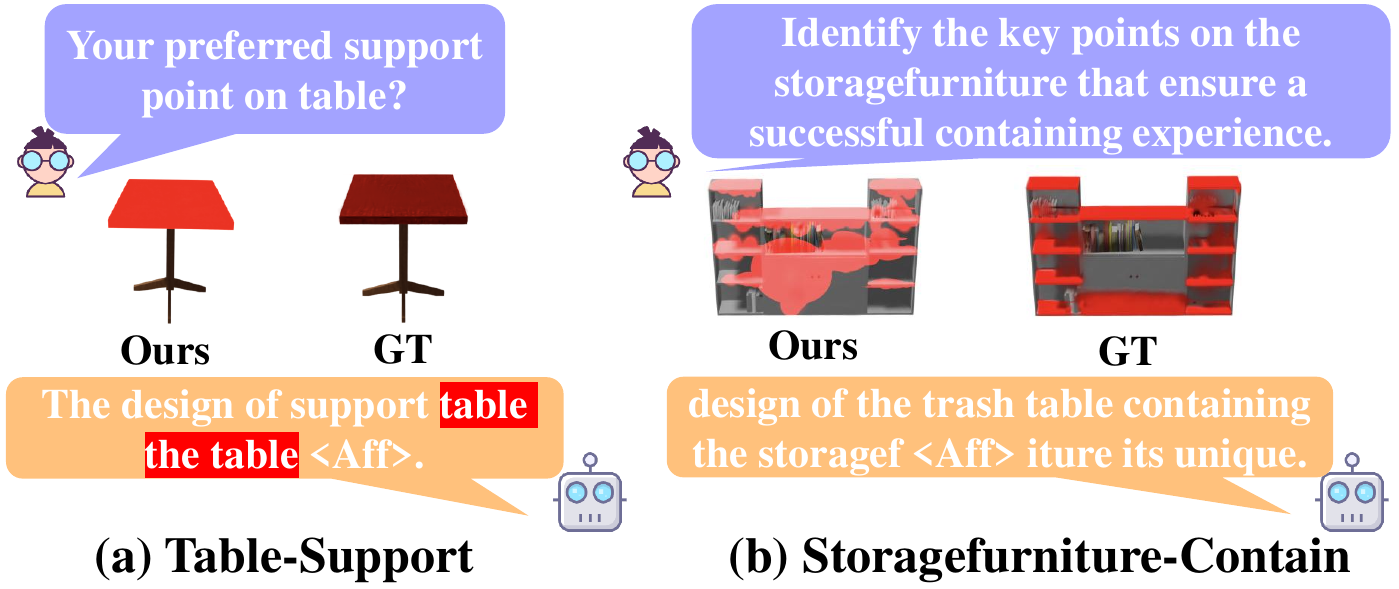}
        \vspace{-10pt}
    \caption{ \textbf{Failure Cases}. (a) Incorrect language response and (b) Insufficient ability to handle complex object architecture.}
        \vspace{-10pt}   
    \label{fig:Failure-cases}
\end{figure}

\noindent{\textbf{Failure Analysis.}} As shown in \autoref{fig:Failure-cases}, the primary causes of failure are incorrect answers and erroneous annotations. The model's performance degrades when processing complex instructions, leading to suboptimal responses. This issue can be attributed to limitations in the language models used, such as RoBERTa~\cite{liu2019roberta}, GPT-2~\cite{lagler2013gpt2}, and BART~\cite{lewis2019bart}, which have smaller parameter sizes and vocabulary coverage insufficient for comprehensive affordance reasoning. Specifically, RoBERTa's~\cite{liu2019roberta} limited vocabulary restricts the model's ability to generate precise text responses, highlighting the need for more advanced language models in future work. Additionally, the model struggles with objects that have multiple discontinuous affordance regions, such as multi-layered storage furniture, further indicating areas for improvement in model architecture and training strategies.


\section{Potential Applications}
\noindent{\textbf{Robotic Task Planning with Geometric-Aware Affordance Reasoning.}}
Recent works like RT-Affordance~\cite{nasiriany2024rtaffordance} and PLATO~\cite{car2024plato} highlight the need for fine-grained affordance understanding to bridge high-level language instructions and precise robotic manipulation. 3DAffordSplat’s high-fidelity Gaussian representations enable robots to identify geometrically accurate interaction regions (e.g., graspable handles, rotatable joints) in cluttered environments, addressing limitations of point cloud-based methods~\cite{li2024laso,yang2023grounding, luo2022learning} in industrial assembly or household tasks. Future integration with LLM-driven planners~\cite{chu2025daffordancellm, qian2024affordancellm} could enable zero-shot adaptation to novel objects, particularly for deformable or articulated items where continuous surface modeling is critical.

\noindent{\textbf{Augmented Reality (AR) Interfaces for Interactive 3D Scene Understanding}} The real-time rendering capability of 3DGS~\cite{kerbl20233d} combined with AffordSplatNet’s affordance reasoning aligns with emerging AR frameworks like LangSplat~\cite{qin2024langsplat} and Feature3DGS~\cite{zhou2024feature}, which require dynamic interaction with 3D scenes. Applications include furniture arrangement assistants that highlight "placeable" surfaces or maintenance training systems visualizing "rotatable" mechanical parts. This could extend to physics-aware AR simulations, leveraging the structural consistency of Gaussian splats to predict interaction outcomes (e.g., door opening trajectories).

\noindent{\textbf{Context-Aware Smart Home Systems}} Building on embodied AI frameworks like MoMa-Kitchen~\cite{zhang2025moma} and AGPIL~\cite{zhu2025grounding3dobjectaffordance}, 3DAffordSplat’s multi-modal alignment enables intelligent environments to interpret user intents through spatial affordances. For example, a voice-activated system could identify "pushable" cabinet doors or "liftable" sofa cushions by correlating language queries with Gaussian-based structural features. Future integration with IoT sensors could enable adaptive interfaces that update affordance predictions based on object state changes (e.g., detecting "unstable" furniture poses after collisions).

\noindent{\textbf{Industrial Quality Control via Cross-Modal Defect Detection}} Recent studies in 3D anomaly detection~\cite{lu2024geal,yu2024seqafford} emphasize the need for robust geometric reasoning in manufacturing. AffordSplatNet’s cross-modal alignment module could identify functional defects (e.g., misaligned "slidable" rail components) by comparing ideal Gaussian affordance maps with LiDAR-scanned point clouds of production-line objects. This aligns with Industry 5.0 trends toward AI-driven preventive maintenance, where deviations from expected affordance patterns (e.g., "non-rotatable" bearings) signal potential failures before physical testing.

{
    \small
    \bibliographystyle{ieeenat_fullname}
    \bibliography{main}
}
\end{document}